\documentclass{ieeeaccess}
\usepackage[square, comma, sort & compress, numbers]{natbib}
\usepackage{amsmath,amssymb,amsfonts}
\usepackage{mathrsfs}
\usepackage{mathtools}
\usepackage{algorithmic}
\usepackage{graphicx}
\usepackage[colorlinks, linkcolor=blue, verbose]{hyperref}
\usepackage{textcomp}
\usepackage[ruled,linesnumbered]{algorithm2e}
\def\BibTeX{{\rm B\kern-.05em{\sc i\kern-.025em b}\kern-.08em
		T\kern-.1667em\lower.7ex\hbox{E}\kern-.125emX}}
\begin{document}
	
	\doi{}
	
\title{A Comparative Analysis of Machine Learning and Grey Models}
\author{\uppercase{GANG HE}\href{https://orcid.org/0000-0002-4029-7596}{\includegraphics[scale=0.10]{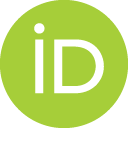}}\authorrefmark{1,*}, 
\uppercase{KHWAJA MUTAHIR AHMAD}\href{https://orcid.org/0000-0002-1115-7690}{\includegraphics[scale=0.10]{orcid.png}}\authorrefmark{2,*}, 
\uppercase{WENXIN YU}\href{https://orcid.org/0000-0002-6093-5516}{\includegraphics[scale=0.10]{orcid.png}}\authorrefmark{1}, 
\uppercase{XIAOCHUAN XU}\href{https://orcid.org/0000-0001-5171-5549}{\includegraphics[scale=0.10]{orcid.png}}\authorrefmark{3}, and
\uppercase{Jay Kumar}\href{https://orcid.org/0000-0003-4915-9701}{\includegraphics[scale=0.10]{orcid.png}}\authorrefmark{2}
	}
	\address[1]{School of Computer Science and Technology, Southwest University of Science and Technology, 621010 Mianyang, Sichuan, P.R.China.}
	\address[2]{School of Computer Science and Engineering, University of Electronic Science and Technology of China, 611731 Chengdu, Sichuan, P.R.China.}
	\address[3]{School of Sciences, Southwest Petroleum University, 610500 Chengdu, Sichuan, P.R.China.}

	\tfootnote{}
	
	\markboth
	{G. He, K. M. Ahmad \headeretal: A Comparative Analysis of Machine Learning and Grey Models}
	{G. He, K. M. Ahmad \headeretal: A Comparative Analysis of Machine Learning and Grey Models}
	
	\corresp{Corresponding author: Gang He (e-mail:ganghe@swust.edu.cn).}

\begin{abstract}
Artificial Intelligence (AI) has recently shown its capabilities for almost every field of life. Machine Learning, which is a subset of AI, is a `hot' topic for researchers. Machine Learning outperforms other classical forecasting techniques in almost all-natural applications and it is a crucial part of modern research. Many modern Machine Learning methods require a large amount of training data. Due to the small datasets, the researchers may not prefer to use Machine Learning algorithms that require large training data. To tackle this issue, this survey illustrates, and demonstrates related studies for significance of Grey Machine Learning (GML). Which is capable of handling large datasets as well as small datasets for time series forecasting likely outcomes. This survey presents a comprehensive overview of the existing grey models and machine learning forecasting techniques. To allow an in-depth understanding for the readers, a brief description of Machine Learning, as well as various forms of conventional grey forecasting models are discussed. Moreover, a brief description on the importance of GML framework is presented.
\end{abstract}


\titlepgskip=-15pt
	\maketitle
\let\thefootnote\relax\footnotetext{*These authors contributed equally to this work.}

\section{Introduction}
\label{sec:introduction}
\PARstart{M}{achine} learning techniques plays an essential role in all fields of social science as well as natural science especially for forecasting \cite{chen2017disease,steffenel2021forecasting}. Every country has a relevant organization that analyzes, and collects the economic facts, and figures to predict future tendency for several economic indicators and assist policy-makers in their decision-making \cite{jia2020similarity,Weigend1994TimeSP}. Data gathered from industries (e.g., demand and sale) remain insufficient. Recently, several types of forecasting methods were proposed and can be divided into two main categories: (i) Qualitative and (ii) Quantitative. Qualitative methods include expert system, trend prediction, Delphi, etc. Meanwhile, the quantitative methods include multi-linear regression analysis, exponential smoothing, time series analysis, and genetic algorithms \cite{ribeiro2020imbalanced,Cerqueira2020EvaluatingTS}. These forecasting methods are constrained by the lack of data, complicated input variables, and predicted environmental changes \cite{hsu2003applying}. There are more than 300 studies related to forecasting. However, only a few of such studies are reliable. Despite the well-developed scientific technologies, There are several social and natural factors which are unexplained, uncertain, or incomplete. Besides the availability of an extensive range of technologies and frameworks, which can be used for big datasets \cite{chen2008application,reddy2020analysis}.

\Figure[t!][width=6.0in,height=6.5in,clip,keepaspectratio]{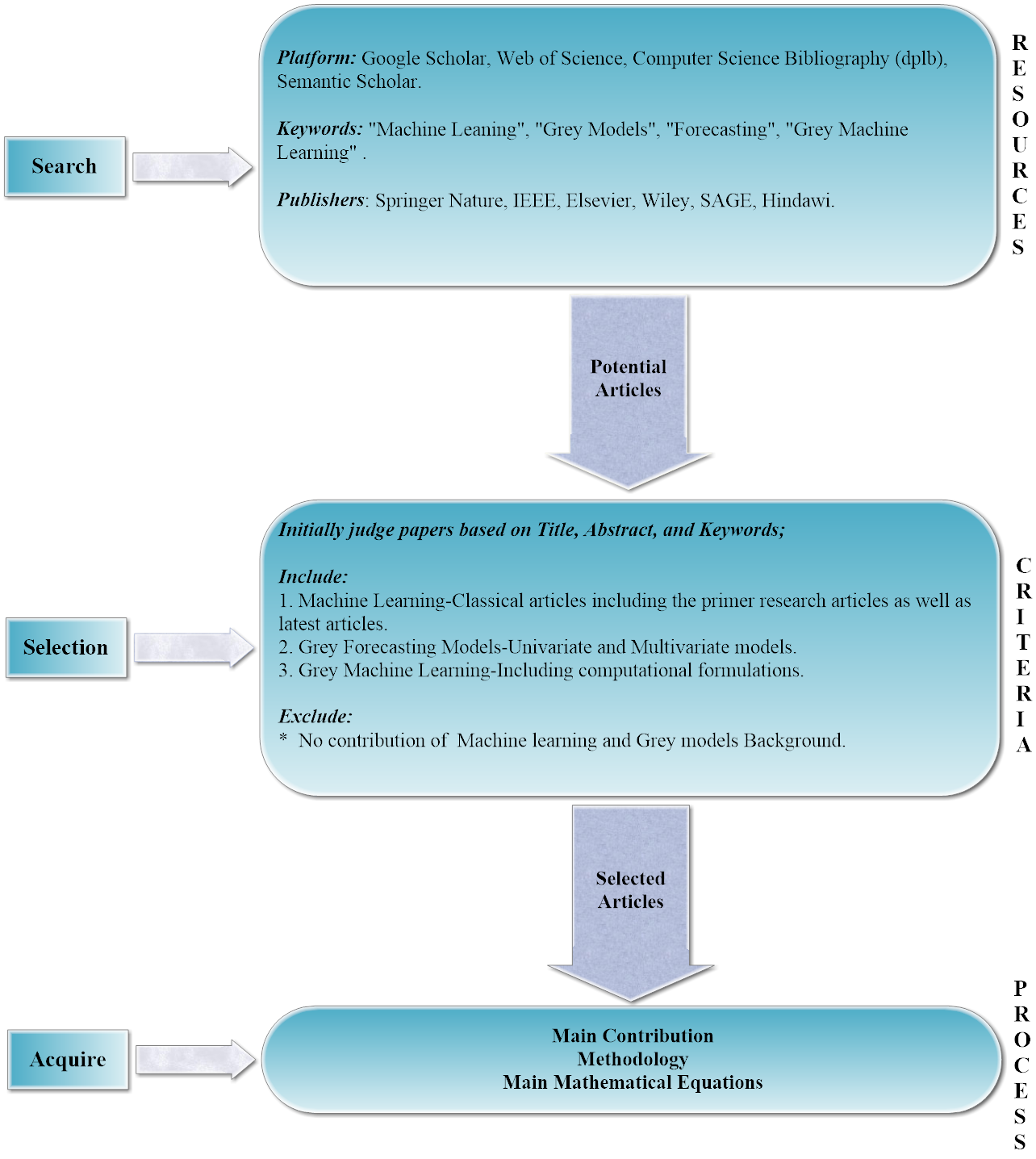}
{Illustration the criteria and selection process of this survey paper within the domain of building optimization.\label{fig1}}

Recent forecasting review studies offer a systematic overview of current forecasting models and their classification. Hippert et al. \cite{hippert2001neural} presented a review on short-term load forecasting. Mat Daut et al. presented a review on building electrical energy consumption forecasting analysis using conventional and AI methods \cite{zimmer2008ten}. Zhao et al. classified and reviewed the existing methods for building energy consumption prediction \cite{zhao2012review}. A review on energy demand models for forecasting proposed by Suganthi and Samuel \cite{suganthi2012energy}. Fumo et al. presented a detailed study on building energy estimation and classification \cite{fumo2014review}. Furthermore, Martinez-Alvarez et al. presented a survey on data mining techniques for time series forecasting of electricity \cite{martinez2015survey}. Raza and Khosravi presented a review on short-term load forecasting techniques based on AI techniques \cite{raza2015khosravia}. Wang et al. proposed a review of AI based building energy prediction with a focus on ensemble prediction models \cite{wang2015review}.

Recently, AI has shown its capabilities for almost every field of life. In 2015, The authors designed a machine to understand Mongolian and passed the Turing test \cite{lake2015human}. Such research has shown that the computer can function as humans in handwriting tasks. In 2016, David Silver et al. published AlphaGo's first paper, stating that the machine could beat the Go game practitioners for the very first time. This article was published after AlphaGo defeated the World Champion Lee Sedol. The success of AlphaGo had also demonstrated that the machine can not only be like humans, but can also be smarter than humans, and the output of this research gives AI enormous confidence \cite{silver2016mastering}. Chirag et al. presented a review on time series forecasting techniques for building energy consumption \cite{deb2017review}. The superior performance of the hybrid and ensemble models for time series forecasting was also proposed in recent review studies \cite{tealab2018time,hajirahimi2019hybrid,prilistya2020tourism,lara2021experimental,xia2021conditional}. All the above surveys and studies provided vital information on forecasting models on different scales.

Besides this, a question that needs to be addressed is how can researchers use Machine Learning techniques using extremely small datasets to acquire high accuracy and speed?

To answer this, a comprehensive overview of Machine Learning, grey models, and GML framework illustrated in this survey to highlight the present outlooks, and enhancement for the classical proposed methods.

\subsection{Objectives of the survey}
A forecasting model can be rely on static data that compares a dependent variable to a collection of independent variables, or it can be rely on composite or simultaneous time series data \cite{deb2017review}. The importance of time series analysis has evolved as people more convinced of the significance of real-time data monitoring and storage \cite{aldhyani2020intelligent}. The aim of this survey is to understand more about the existing time series forecasting framework known as GML. The key contributions of this article can be summarized as follows:
\begin{itemize}
\item First, a comprehensive survey on evolving and futuristic Machine Learning techniques is presented. In particular, a primer overview (1950-2021) of algorithms, and emerging applications is presented to highlight the present outlooks, and enhancement for the classical proposed methods.
\item Second, the formulation of conventional grey forecasting models is presented including an overview. Moreover, both univariate and multivariate Grey Models are discussed in detail. Particular emphasis is given to the mathematical background.
\item Finally, the general formulation for black-box learning and white-box learning framework called GML is presented. For brevity, only the key derivations and computational formulations are discussed.
\end{itemize}
This subsection describes the process of this paper and provides a summary of the articles discussed in Figure \ref{fig1}. The structure of this survey is given as follows: To illustrate the core concept of GML. Firstly, we provide a primer survey of Machine Learning algorithms and their applications in Section \ref{sec:Machine Learning}. The general forms of the conventional grey models, including popular models, are illustrated in Section \ref{sec:Conventional Grey Models}. The general formulations including the computational details of GML are discussed in Section \ref{sec:GML}. A brief discussion and future perspectives of GML have been presented in Section \ref{sec:dis}. Finally, the conclusion is presented in Section \ref{sec:conclusion}.

\section{Machine Learning}
\label{sec:Machine Learning}
Machine Learning is a subset of AI specifically developed to simulate human intelligence. It is essentially a data processing approach that automates the construction of an empirical model. In other words, it is focus on the premise that algorithms can learn from data, recognize patterns and make decisions with minimal human intervention. Recently, Machine Learning techniques have been used for Big Data \cite{travis2021kernel} and has been widely implemented in some of the areas, varying from computer vision \cite{viejo2018robotics}, finance \cite{lee2019global}, spacecraft engineering \cite{lucas2017machine}, entertainment \cite{vcerticky2019psychophysiological}, pattern recognition \cite{dwivedi2016software}, and computational biology to biomedical applications \cite{Baloglu2021WhatIM}. In this overview, the types of Machine Learning algorithms as well as popular techniques of Machine Learning are presented.
\subsection{Overview of Machine Learning}
In this subsection, a brief analysis over time to explore the history of Machine Learning as well as the most significant milestones are presented. Furthermore, we have divided this overview into several categories to make it more understandable.

\subsubsection{Classical work}
Although Arthur Samuel et al. was an American pioneer of AI and inventor of the term ``Machine Learning'' in International Business Machines (IBM) a leading US computer manufacturer in 1959 \cite{kohavi1998glossary}, the year 1950 marks the first time Alan Turing et al. proposed the ``Turing Machine'' while posing such questions as, ``Can machines think?'' and ``Can machines do what we can do? (as thinking entities)''. For instance, if a machine has true intellect, then the machine device must have the ability to trick a human being into thinking that it is also a human \cite{harnad2006annotation}. In Turings' proposed study, there are many features that could be exhibited by machine intelligence and the different consequences for architecture are revealed, this is the first discovery in the field of Machine Learning. Earlier research works planned to connect the computer with human interaction. For this purpose, the first Machine Learning program was published in 1953, which is written by Arthur Samuel et al. \cite{kohavi1998glossary}. The software was a game (Checkers). The IBM machine improved the design of the game and its progress, helped to refine the winning tactics, and integrated certain movements into the software. Based on the above studies, it can be analyzed that, in the early stages of Machine Learning developments, scientists and engineers introduced Machine Learning applications to improve computer intelligence.

\Figure[t!][width=6.8in,height=7.5in,clip,keepaspectratio]{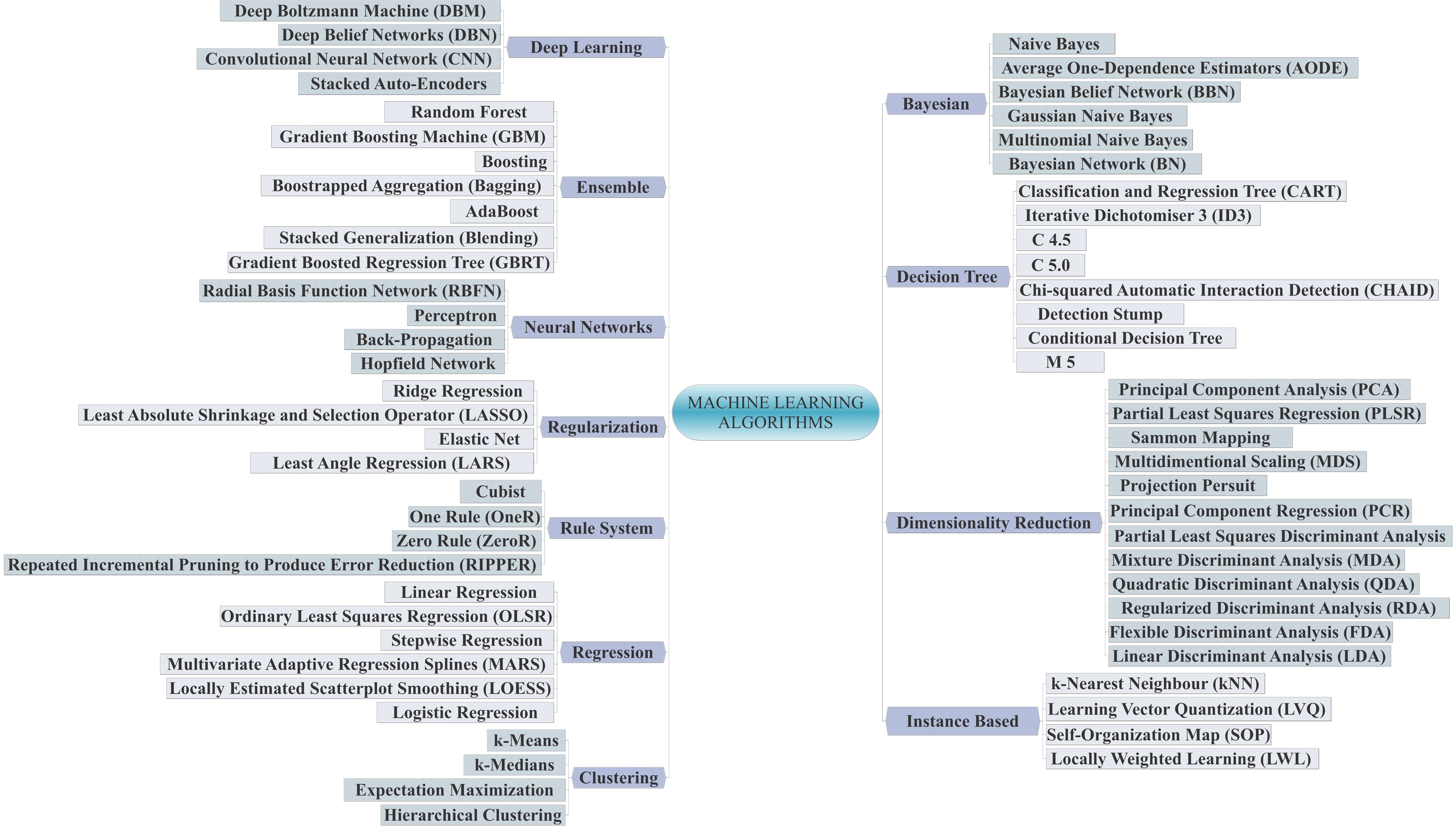}
{Popular Machine Learning Algorithms.\label{fig2}}

In 1957, Frank Rosenblatt et al. proposed the first neural network for computers namely called ``the perceptron'' \cite{rosenblatt1958perceptron}, which simulates the thinking patterns in the human brain. Moreover, T.M. Cover and P.E. Hart proposed the ``nearest neighbor'' algorithm in 1967 \cite{cover1967nearest}. This kind of algorithm is used for simple pattern recognition. Essentially, It was used to make a path for passengers beginning with a random location, but to make sure that they reach all cities on a short ride. Moreover, Stanford University students developed the ``Stanford Cart'' in 1979 \cite{moravec1980obstacle}, which can navigate obstacles automatically. The Cart was a moderately remotely controlled television-equipped mobile robot in the Stanford Artificial Intelligence Laboratory (SAIL). Other researchers proposed several explanation-based approaches, which can be linked back to the MACROPS learning strategies used in STRIPS \cite{fikes1972learning}. The key sheet of Machine Learning algorithms have been discussed in Figure \ref{fig2}.

According to the expansion and enhancement, it must be credited to Silver, Mitchell, and DeJong. At the same period, each of these researchers built very broad Explanation Based Learning (EBL) computer systems-LP (Silver), ESA (DeJong), and LEX2 (Mitchell). This kind of systems analyzes training data and generates a basic law that can be enforced by discarding the redundant data. It presents a historical account of the development of EBL and discusses some of the important outstanding research tasks \cite{dejong1986explanation}. Due to the increasing size of data, in 1998, the Scientists attracted a large number of data programs and applications (Data-Driven Approach using Machine Learning) \cite{pirrelli1999hidden,wang1999computer}.

The above history shows that many researchers have been doing their research from time to time and played a key role in the field of Machine Learning. In the 21st century, the new millennium introduced an abundance of integrated technology. There is Machine Learning where adaptive programs are expected. These algorithms are capable of detecting patterns, extracting new knowledge from inputs, learning from practice, and optimizing the efficiency and accuracy of their analysis and output. Essentially, the researchers have taken the main ideas from the earlier proposed works and have enhanced their work in the form of new inventions to develop new algorithms, software, games, as well as probabilistic reasoning, particularly in the field of automated medical diagnosis \cite{stuart2003artificial}.

In 2000, Thomas G. Dietterich et al. \cite{dietterich2000ensemble} proposed the ensemble methods in Machine Learning. As we know the original set approach is Bayesian averaging, but several techniques offer error-correcting output coding, boosting, and bagging. The authors analyzed these approaches and clarified why the ensembles can often do better than any single classifier \cite{schapire2003boosting}. In 2002, the authors designed a new technique for gene selection using Support Vector Machine (SVM) approaches focused on Recursive Feature Elimination (RFE) for cancer classification \cite{guyon2002gene}. Experimentally, they proved that the genes identified by their strategies yield better detection efficiency and are biologically important to cancer. This research shows that Machine Learning models have already been used to diagnose human diseases. In other research, the authors suggested that Machine Learning should also be a faster and better approach to corner-detection \cite{rosten2008faster}.

\Figure[t!][width=6.8in,height=8in,clip,keepaspectratio]{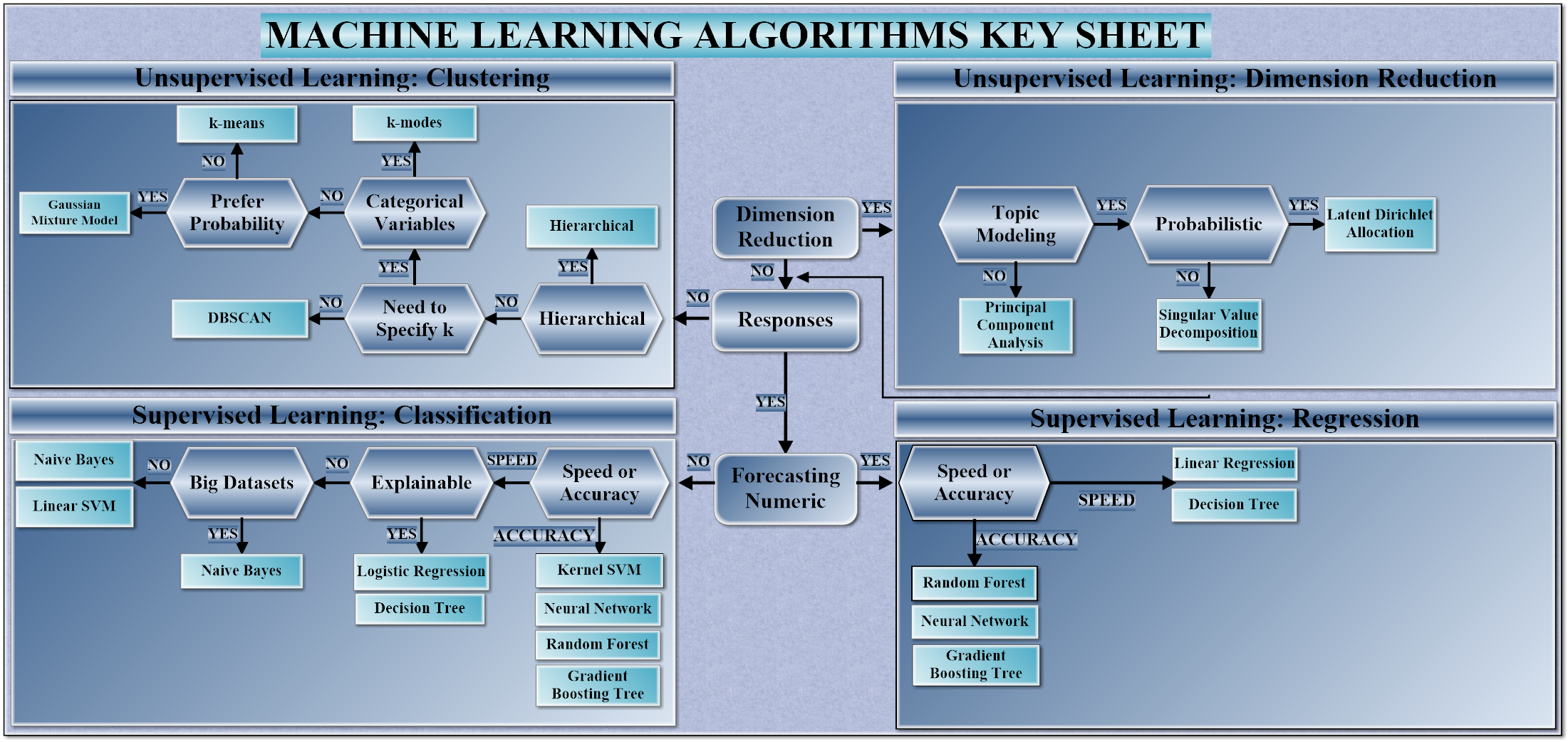}
{Key sheet of Machine Learning Algorithms.\label{fig3}}

\subsubsection{\textbf{State-of-the-art}}
Recently, several researchers focusing on Machine Learning technologies and algorithms merge diverse areas to increase production efficiencies, such as cloud computing, biomedical engineering, network security, image processing, forecasting, Internet of Things (IoT), and Big Data technology \cite{zhu2021iot,pelissier2020combining,2019Tunnel,Naeem2020VehicleTE,Yan2020Short,raza2020comprehensive}. In 2019, Wenrui Yang et al. introduced a system for sports image detection using Machine Learning. The purpose of this study was to develop the identification of athletes, the judgment on sport behavior, the perception of motion, and the development of a test framework to validate the effectiveness of the research process \cite{yang2019analysis}. In another study, the authors suggested a combination of big data processing using cloud computing and Machine Learning \cite{sharma2019big}. In this study, they proposed that, Machine Learning seems to be an ideal solution for exploring the possibilities concealed for big data. Recently, the authors published a performance analysis of the new Machine Learning algorithm and the Logistic Prediction Method called MLIA \cite{ma2019financial}. In this review, the authors demonstrated that Machine Learning has a significant predictive effect of MLIA on the measurement of financial credit risk and can provide a theoretical basis for subsequent relevant studies.

In addition to the role of Machine Learning in wireless communication, Machine Learning techniques are anticipated to play a significant part in the implementation of the fifth-generation (5G). Researchers have recently presented an overview of wireless communication channel modeling based on Machine Learning \cite{Aldossari2019MachineLF}. In this overview, the writers addressed 5G with massive Multiple-Input / Multiple-Output (MIMO), quick handover, higher data rate, and channel simulation becoming more complicated than other conventional stochastic or deterministic models. To this purpose, scholars and academics are looking forward to more effective methods that are less complicated and more reliable. For example, Emerging Machine Learning Methods can offer a new direction for the analysis of big data and traffic data. To diagnosing human diseases by using Deep learning techniques \cite{hernandez2021machine}  such as, the Convolutional Neural Networks (CNNs) have provided significant performance to boost the fields related to diagnosing human diseases. CNN techniques have been applied successfully for several tasks, like computer-aided diagnosis, image enhancement/generation, classification, and segmentation \cite{Pei2017NonrigidC2,huang2017lung,oktay2017anatomically,gouk2021regularisation,cheema2019liver,shakeel2021detecting}. In comparison of Machine Learning with IoT and Big Data, Jangam J.S Mani and Sandhya Rani Kasireddy propounded a framework that classifies the population into four classes based on diet efficiency. After 30 days of dietary retrieval, they devour as normal, unbalanced, almost balanced, and almost unbalanced by using logistic regression, linear discriminant analysis (LDA), and random forest algorithm \cite{mani2019population}. In Figure \ref{fig3}, the popular Machine Learning algorithms have been demonstrated.

To discuss the data processing architecture, Machine Learning approaches are used to allow precise tuning to train a classifier for large-scale datasets. To serve IoT applications, these infrastructures use Machine Learning or AI-based techniques which evaluate entity or system data to produce valuable knowledge that can be used for service or decision-making. Machine Learning methods make it possible for machines to communicate with people, drive cars automatically, forecasting, writing and publishing sport match reports, and identifying criminal suspects as well. Furthermore, Machine Learning has a serious impact on most businesses and employees inside them, that is why a professional will at least have a context about what Machine Learning is, and how it is evolving  \cite{khan2019machine,steffenel2021forecasting,alam2018survey}.

\Figure[t!][width=6.5in,height=7in,clip,keepaspectratio]{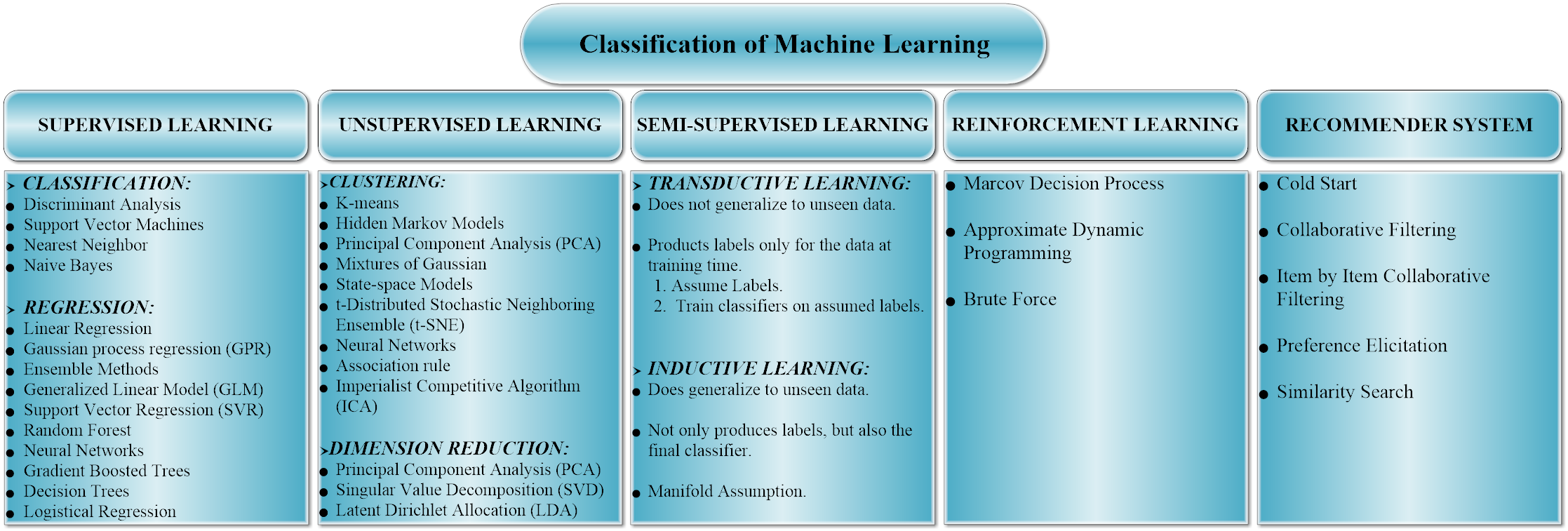}
{Classification of Machine Learning Algorithms.\label{fig4}}

\subsection{\textbf{Popular Algorithms of Machine Learning}}
In this subsection, popular algorithms of Machine Learning are presented. Machine Learning algorithms are specifically programmed to create predictive models dependent on the underlying algorithm and dataset. Input data for Machine Learning algorithms usually consist of  ``label'' and ``features'' over a range of samples. Labels are what the purpose of a Machine Learning algorithm is to determine, which is the output of the model, whereas the features are the quantities of all tests, either raw or mathematically transformed \cite{camacho2018next}. The most common Machine Learning algorithms can be divided into two key categories-Supervised learning and Unsupervised learning \cite{rencher2005review,James2013AnIT}. Apart from these, some other methodologies of Machine Learning are also discussed in Figure \ref{fig4}.

\subsubsection{\textbf{Supervised Machine Learning Algorithms}}
Supervised Machine Learning is the cognitive activity of discovering relationships between parameters in annotated data (training set). Using this knowledge, making a forecasting model capable of inferring annotations for new data in which annotations are unknown. This kind of algorithm uses the characteristics and annotations of the training set to induce the model to predict the annotations of instances in the test set \cite{fabris2017review,8862225}.

\subsubsection{\textbf{Unsupervised Machine Learning Algorithms}}
Compared with Supervised learning algorithms, Unsupervised learning algorithms work without the desired output label. For instance, an unsupervised learning algorithm analyzes the $a$ without needing the $b$, whereas a Supervised Machine Learning algorithm usually learns from a method that maps an input $a$ into the $b$ output. Unsupervised learning strategies may be motivated by theoretical and Bayesian concepts of intelligence. Unsupervised learning algorithms usually using to expand the data and train a model for finding suitable internal representation, such as sorting data into clusters \cite{khanum2015survey,bertsimas2021interpretable}.

\subsubsection{\textbf{Semi-supervised Machine Learning Algorithms}}
Traditional classifiers need to train labeled data (features/label pairs). However, labeled instances are sometimes challenging to procure, time taking and costly, since they involve the efforts of the skilled human annotator. In the meantime, unlabeled data can be relatively easy to get but difficult to use. Semi-supervised learning solves this issue by creating stronger classifiers utilizing vast volumes of unlabeled data, coupled with the labeled data needing fewer human intervention and higher accuracy \cite{zhu2005semi,bekker2020learning}. Semi-supervised learning algorithms are used in such cases where the labels are missing. For instance, only a limited amount of training data is labeled and the goal is to improve the output of the model that can be done either by avoiding the labels and performing unsupervised learning or by ignoring unlabeled data and performing supervised learning. It is the great interest in both practical and theoretical aspects \cite{camacho2018next,lucas2021bayesian,wu2021semi}.

\subsubsection{\textbf{Reinforcement Machine Learning Algorithms}}
Reinforcement learning is a branch of Machine Learning algorithms in which the learner or software entity tries to perform a sequence of acts that will optimize accumulated incentives, such as winning a checker or chess game. It is an area of research that has been able to overcome a broad variety of complicated decision-making problems which were historically been out of control for the machine. It also opens up a range of new opportunities in areas such as infrastructure, automation, smart grids, banking, and much more \cite{sutton1998introduction,franccois2018introduction,zhang2021learning}.

These methods led to impressive advances in AI, going beyond human performance in domains ranging from Atari to Go to no-limit poker \cite{Wang2019BeyondWA}. These signs of progress attracted the attention of cognitive scientists interested in understanding human learning. Over the last few years, because of its performance in solving the complexities of sequential decision-making, it has become increasingly popular. Some of the achievements were attributed to the combination of reinforcement learning and deep learning methodologies \cite{lecun2015deep,goodfellow2016deep,schmidhuber2015deep,botvinick2019reinforcement}.

\subsubsection{\textbf{Recommender Systems}}
Recommender systems have been built in coexistence with the internet. Initially, this kind of systems were focused on statistical, content-based and shared filtering. Such systems currently integrate social knowledge. Recommender systems can also be described as learning techniques through which online customers can design their websites to match the customer's tastes such as, an internet customer may get a product and/or associated products ranking while looking for things based on an established recommendation system. There are mainly two methods, namely content-based recommendation, and collective recommendation. This type of system allows users to get access to it and collect info, principles, intelligent and novel suggestions. Several e-commerce pages use this program \cite{bobadilla2013recommender,Ali2021,himeur2021survey,ali2020context}.
\Figure[t!][width=6in,height=6.5in,clip,keepaspectratio]{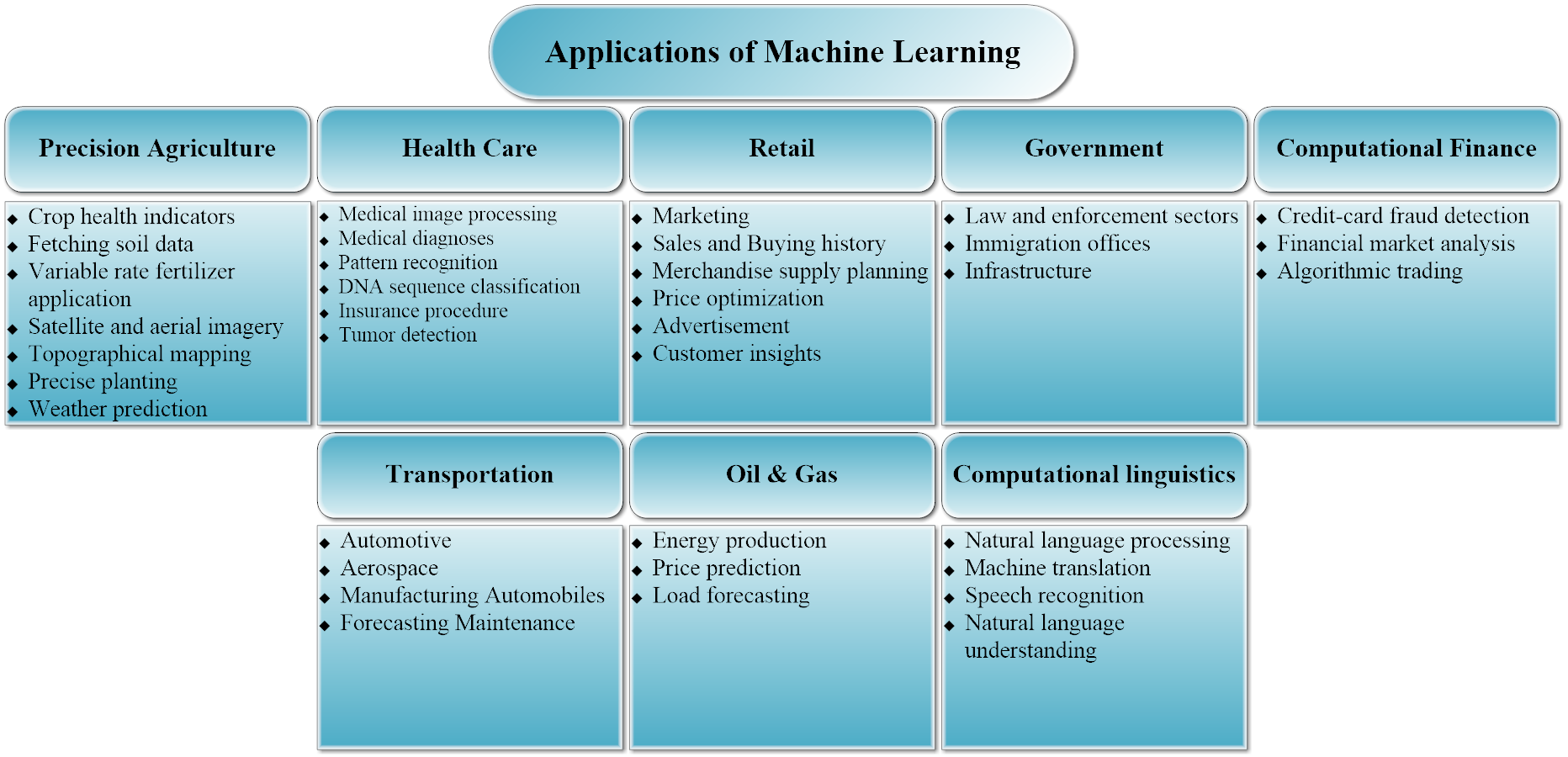}
{Popular Applications of Machine Learning.\label{fig5}}
\subsection{\textbf{Popular Applications of Machine Learning}}
Many companies and industries dealing with data packages have recognized the importance of Machine Learning technology. By using Machine Learning approaches, businesses can work more reliably and effectively as well as gain an advantage over competitors \cite{ccavucsouglu2019new,hullermeier2019aleatoric}.

Furthermore, with the aid of compelling articles for clarity to the reader, several applications of Machine Learning are also discussed and divided into different sections in Figure \ref{fig5}, which are given below:

\subsubsection{\textbf{Precision Agriculture}}
The most common concepts of Machine Learning in the field of agriculture were proposed by  A. Kukuta et al. \cite{kukutai2016can}. Precision agriculture, satellite farming, or site-specific crop management are agricultural management terms focused on observation, estimation, and the reaction of inter-and intra-field crop variability. The key goal of precision agriculture analysis is to set up a decision support system for agricultural management to optimize the return on inputs while maintaining energy \cite{whelan2003definition,mcbratney2005future,milella2019multi}.

\subsubsection{\textbf{Health care}}
Machine Learning is an emerging field and fast-growing phenomenon in the field of health care \cite{van2021artificial,he2010integrated}. The advent of smart applications and devices that can use data to assess the health of patients in real-time. The medical professionals analyze data for the detection of patterns or warning flags that can contribute to better diagnosis and care \cite{beam2018big,char2018implementing,ngiam2019big,Chand2020Two}.

\subsubsection{\textbf{Retail}}
Websites recommend items that you would buy based on prior purchases using a Machine Learning method called a `recommendation system' to evaluate your experience about purchasing the item. Retailers depend on Machine Learning technology to record, interpret, and customize their shopping experience \cite{aguilar2020cold,gately2017vekia}.

\subsubsection{\textbf{Government}}
State departments, such as infrastructure and public health have a strong need for artificial intelligence because they offer several data points that can be exploited for information. For instance, analyzing the data of the system to find opportunities to boost efficiency and save money. Machine Learning can also help spot fraud and prevent data theft \cite{piscopo2017predicting,Alexopoulos2019HowML}.

\subsubsection{\textbf{Computational Finance}}
Banks and certain organizations in the business industry are utilizing Machine Learning technologies for two key purposes: the first is to identify valuable insights from data and the second is to prevent fraud. Insights can recognize investment opportunities and allow shareholders to know when to sell. Data mining can also recognize high-risk clients or use cyber monitoring to detect warning signals of fraud \cite{gerlein2016evaluating,Prado2018AdvancesIF}.

\subsubsection{\textbf{Transportation}}
Analyzing data to detect patterns and developments is crucial for the transport sector, which focuses on keeping roads more effective and finding possible challenges to improve productivity. Information modeling and simulation elements of Machine Learning are useful tools for logistics firms, urban transportation's, and other transit organizations \cite{jahangiri2015applying,tizghadam2019machine}.

\subsubsection{\textbf{Oil and gas}}
Use cases include finding new sources of energy, analyzing elements in rocks, predicting malfunction of refinery sensor, streamlining the production of oil to make it more reliable and cost-effective. The amount of use cases of Machine Learning in this sector is overwhelming and continues to increase \cite{mohamed2015machine,hajizadeh2019machine,2019Yan,2020A}.

\subsubsection{\textbf{Computational linguistics}}
Computational linguistics has historically been conducted by computer scientists who have specialized in the use of computers for the analysis of natural languages. Nowadays, computer linguists often work as part of interdisciplinary teams, which can include computer scientists, target language specialists, and professional linguists \cite{amodei2016deep,alzantot2018did}.

\section{\textbf{Conventional Grey Models}}
\label{sec:Conventional Grey Models}

Basic Grey Model Theory is an interdisciplinary research discipline that was proposed by \cite{deng1982control} in 1980. He provided a classic continuous GM(1,1) model in which procedures begin with a differential equation namely `whitening equation'. As long as knowledge is concerned, systems that suffer from a lack of information, such as operating mechanism, structure message, and actions log, are referred to as Grey Systems. Throughout the background, the human body, livestock, climate, etc., are the Grey Systems, where ``grey'' implies incomplete, unknown, poor, etc. The goal of the Grey Model and its applications is to bridge the distance between natural science and social science \cite{yu2020n}. This concept has been very popular in terms of its potential to work with systems that have partly unknown parameters. As an improvement over traditional predictive models, grey forecasting models only need small datasets to determine the actions of unknown processes \cite{Julong1989Introduction}. Some of the existing grey models with a continuous whitening function follow the same linear formula as follows: 

\begin{align}
\frac{d X^{(1)}(t)}{d t}+a x_{1}^{(1)}(t)=f(\boldsymbol{\theta} ; t),
\end{align}
while the series $X_{1}^{(1)}$ is usually referred to as an input series. Sometimes, the function $f(\boldsymbol{\theta} ; t)$ differs by time $t$ or dependency sequence (or input series) $X_{i}^{(1)}, i=2,3, \dots, n$, with unknown parameters $\theta$.
In case of discrete grey models, the general linear equation can also be written as,
\begin{align}
x_{1}^{(1)}(k+1)=\alpha x_{1}^{(1)}(k)+f(\boldsymbol{\theta} ; k).
\end{align}
It is clear that the solutions of grey models in the above-mentioned formulations also contain similar formulations and the same equations. Several grey models also use the initial condition $X_{1}^{(1)}(1)=X_{1}^{(0)}(1)$. The general formulation of these models can thus easily be accessed.\\
For continuous models, the answer can always be described as the following convolution equation:
\begin{align}
\hat{X}_{1}^{(1)}(t)=X_{1}^{(0)}(1) \cdot e^{-a(t-1)}+\int_{1}^{t} e^{-a(t-\tau)} f(\boldsymbol{\theta} ; \tau) d \tau.
\label{eq:3}
\end{align}
Whereas, in the case of discrete models, the solution can always be described as the following discrete convolution equation: 
\begin{align}
\hat{X}_{1}^{(1)}(k+1)=X_{1}^{(0)}(1) \cdot \alpha^{k}+\sum_{\tau=2}^{k+1} \alpha^{(k+1-\tau)} f(\boldsymbol{\theta} ; \tau).
\label{eq:4}
\end{align}

\Figure[t!][width=4.5in,height=6in,clip,keepaspectratio]{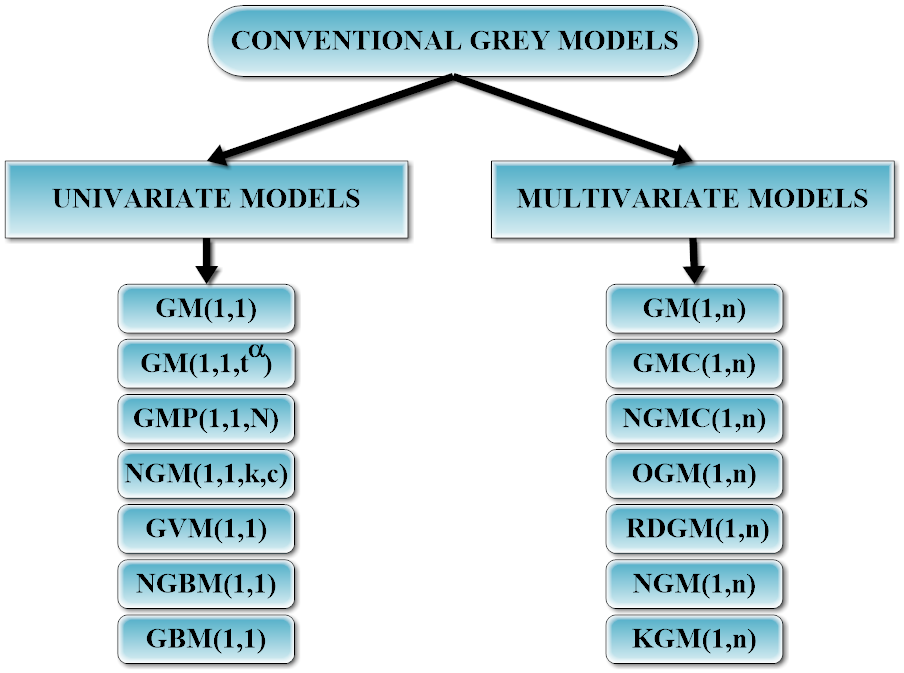}
{Popular Grey Forecasting Models.\label{fig6}}

Grey forecasting models are essentially divided into two categories: Univariate and Multivariate. Single-variable models are called univariate while multivariable models are called multivariate \cite{Liu2010GreyST}. For descriptive purposes, the whitening equation, time response function, and the restored values of grey models are provided in this section and also summarized in Figure \ref{fig6}.

\subsection{\textbf{Evolution of Grey Forecasting Models}}
Since the study of Lin and Liu, the propagation of this Grey System theory took place as follows: Scholarly periodical for the presentation of research results followed by `Journal of Grey System' that started to be published in England in 1989. More than 300 various scientific journals recognize and publish papers relevant to the grey system in the world. Moreover, In the early 1990s, several universities based throughout  China, Taiwan, Australia, United States, and Japan  began offering grey system theory courses. The Chinese Grey System Association (CGSA) was founded in 1996. Every year, CGSA conducts a conference on Grey System Theory and its application \cite{lin2004historical}. In the last four decades, the Grey System Theory has developed quickly and drawn the interest of several researchers. It has been widely and effectively extended to many applications such as commercial, manufacturing, transport, medical, military, mechanical, meteorological, civil, political, financial, science and technology, agricultural, hydrological, geological, etc. Furthermore, the conventional grey model called GM(1,1) has been widely adopted, and its forecasting efficiency could also be improved. To date, several researchers have proposed new approaches for improving the performance of the model as Deng et al. \cite{deng2002basis} have proposed the modifiable residual sequence method. Whereas, Mu et al. \cite{mu2003unbiased} obtained the formula of optimum grey derivative whitening values. He proposed an unbiased GM(1,1) model to develop a framework for estimating the parameters.

Furthermore, Song and Wang developed the center approach of the alteration of grey model GM(1,1). They designed an adjusting grey model \cite{song2001center,wang2001gm}. Tan et al. supported the structure method of the background values in the GM(1,1) model and a basic approximation of the background value function was re-established, which had strong adaptability \cite{tan2000structure}. In another studies, the authors presented the optimal time-response sequence formula and used the least square approach to measure the constant number in the time-response series of the basic GM(1,1) model \cite{tan2000structure,zhang2001accurate,luo2003optimization}. Several researchers examined the appropriate scope and simulation accuracy of the GM(1,1) model \cite{liu2000range,2020SARIMA}. Moreover, some key approaches shall include center approach method \cite{song2001center}, discrete models \cite{xie2005discrete,xie2006research,yao2007improvement}, correcting the residues \cite{deng2002basis}, constructing background values \cite{tan2000structure}, and optimization of the grey derivative \cite{wang2001gm}. A part of these, some other grey forecasting theory methods proposed by scholars \cite{chen2014foundation,qian2012grey,wei2018optimal,chen2010forecasting,zhongqiu1996dynamic,wang2017nls,jie2011grey,ker2011using,wang2014gm,chen2011applying,hao2013piecewise,2020Forecasting}.

\subsubsection{\textbf{Univariate Grey Model}}
The single parameter grey prediction model is GM(1,1) with one vector and one first-order equation, and a simulation unit is a single time sequence. It forecasts the future of the program by defining machine operating rules contained in a series focused on grey generation approaches. The single-variable grey model might not take into account the effect of related factors on the system and thus, it has the benefit of the basic modeling process \cite{zeng2016development}.

Classic GM(1,1) model is an effective method for precise predictions for small samples. Therefore, it is not surprising that GM(1,1) is commonly used as a predictive tool \cite{Liu2010GreyST}. In order to define the GM(1,1) model, the first `1' represents for the `first order', while the second `1' stands the `univariate' \cite{ma2017application}. The system parameters are calculated by de-escalating the whitening equation and using the least square method. For instance, The definition of the Grey Theory is the `Grey Box' where information is established and knowledge is uncertain. Grey system theory is an important tool for determining unknown issues with limited samples and incomplete information \cite{deng1982control}. The Whitening equations of the univariate grey forecasting models have been addressed in Table \ref{tab1}.

\subsubsection*{1. GM(1,1)}
Among the families of grey models, the GM(1,1) model is the most widely used, due to its simplicity and high accuracy for limited datasets \cite{kayacan2010grey}. Based on this essential function, it has been successfully implemented in several fields. Widely used applications of GM (1,1) include energy production \cite{zeng2018forecasting,wang2017grey,zhou2013generalized,wang2017decomposition}, the prediction of stock price \cite{chen2010forecasting},  the oil production in China \cite{ma2018predicting,ma2018kernel}, the consumption of energy \cite{wu2018application,ma2017gmc}, detection \cite{han2020sensor}, and the electricity consumption \cite{wu2018using,zeng2018forecasting}.
Let $X^{(0)}=\left\{x^{(0)}(1), x^{(0)}(2), \ldots, x^{(0)}(n)\right\}$ denote original data, $X^{(1)}=\left\{x^{(1)}(1), x^{(1)}(2), \ldots, x^{(1)}(n)\right\}$ is the first order accumulation generator, $Z^{(1)}=\left\{z^{(1)}(1), z^{(1)}(2), \ldots, z^{(1)}(n)\right\}$ is the background of $X^{(0)}$, where $z^{(1)}(k)=0.5\left(x^{(1)}(k)+x^{(1)}(k-1)\right)$. Hence,
\begin{align}x^{(0)}(k)+a z^{(1)}(k)=b\end{align}
is known as Grey model GM(1,1). The restrictions of $-a$ and $b$ in the grey basic form of GM(1,1) model are referred to as development coefficient and grey action quantity, respectively.
Whereas the time response signal of GM(1,1) is,
\begin{align}
\hat{x}^{(1)}(k+1)=\left(x^{(0)}(1)-\frac{b}{a}\right) \mathrm{e}^{-a k}+\frac{b}{a},   \quad k=1,2, \ldots, n
\end{align}
while the predictive value of GM(1,1) is obtained as,
\begin{align}
&\hat{x}^{(0)}(k+1)=\hat{x}^{(1)}(k+1)-\hat{x}^{(1)}(k)\nonumber\\
&\hspace{18mm}=\left(1-\mathrm{e}^{a}\right)\left(x^{(0)}(1)-\frac{b}{a}\right) \mathrm{e}^{-a k},\nonumber\\
&\hspace{3.5cm} \quad k=1,2, \ldots, n
\end{align}
Let $\widehat{X}^{(0)}=\left\{\hat{x}^{(0)}(1), \hat{x}^{(0)}(2), \ldots, \hat{x}^{(0)}(n)\right\}$. Therefore, ${X}^{(0)}$ is the simulation sequence and $X^{(0)} \cdot \hat{x}^{(1)}(k+1)$ is the simulation data of $x^{(1)}(k+1)$. According to the (3), it is simple to show that the simulation data sequence is the geometric series \cite{lin2004historical}, and thus, the growth rate of the simulation series is constant: 
\begin{align}
&\hat{u}(k)=\dfrac{\hat{x}^{(0)}(k+1)-\hat{x}^{(0)}(k)}{\hat{x}^{(0)}(k)}\nonumber\\
&\hspace{4cm}=\dfrac{\hat{x}^{(0)}(k+1)}{\hat{x}^{(0)}(k)}-1=\mathrm{e}^{ -a}-1.
\end{align}

\subsubsection*{2.NGM(1,1,$k$,$c$)}
Chen and Yu designed a parameter optimization approach to boost the NGM(1, 1, $k$, $c$) model combine with grey action quantity $bt + c$ \cite{chen2014foundation}. Based on the specified form of the latest NGM (1,1,$K$,$c$) model, the differential equation for the NGM(1,1,$k$,$c$) is obtained as,
\begin{align}
\frac{\mathrm{d} x^{(1)}(t)}{\mathrm{d} t}+a x^{(1)}(t)=b t+c,
\end{align}
while $a$ is the developing coefficient, and $bt+c$ is a grey action quantity. Furthermore, the time response function for the NGM(1,1,$k$,$c$) is,
\begin{align}
\hat{x}^{(1)}(k)=\left(x^{(0)}(1)+\frac{b}{a^{2}}-\frac{b}{a}-\frac{c}{a}\right)e^{-a(k-1)}\nonumber\\
+\frac{b}{a} k-\frac{b}{a^{2}}+\frac{c}{a},
k=2,3, \ldots, n,
\end{align}
Whereas the restored function of NGM(1,1,$k$,$c$) model can be written as,
\begin{align}
\hat{x}^{(0)}(k)=\left(x^{(0)}(1)+\frac{b}{a^{2}}-\frac{b}{a}-\frac{c}{a}\right)\nonumber\\
\hspace{2cm}\left(1-e^{a}\right) e^{-a(k-1)}+\frac{b}{a}, \nonumber\\
k=2,3, \ldots, n.
\end{align}

\subsubsection*{3. GM(1,1,${{t}^{\alpha}}$)}
In 2012, Qian et al. developed a new forecasting grey model called GM(1,1,${{t}^{\alpha}}$) with grey action quantity of $b t^{\alpha}+c$, and used it to forecast the settlement of the foundation \cite{qian2012grey}. The whitening differential equation of GM(1, 1,${{t}^{\alpha}}$) model based on \cite{wu2019analysis} is,
\begin{align}
\frac{d x^{(1)}(t)}{d t}+a x^{(1)}(t)=b t^{\alpha}+c, r>0, \alpha>0,
\end{align}
Whereas the time response function of GM(1, 1,${{t}^{\alpha}}$) model is,
\begin{align}
x^{(r)}(k)=\left(x^{(0)}(1)-\frac{c}{a}\right) e^{-a(k-1)}+\frac{c}{a} +\frac{b}{2} e^{-a(k-1)}\nonumber\\
\sum_{-1}^{k-1}\left(\tau^{\alpha} e^{a(\tau-1)}+(\tau+1)^{\alpha} e^{a \tau}\right), \nonumber\\
k=2,3, \ldots, n,
\end{align}
and restored value of $\hat{x}^{(0)}(k)k=2,3, \ldots, n$ is given by,
\begin{align}
x^{(0)}(k)=x^{(1)}(k)-x^{(1)}(k-1).
\end{align}

\begin{table*}[!htbp]
	\caption{The definition of the comparative Univariate grey models used.}
	\renewcommand{\baselinestretch}{2.5}
	{\footnotesize\centerline{\tabcolsep=6pt
			\begin{tabular}{cccccccccccccc}
				\hline
				No.     &    Model Name           &   Differential Equation    &  Reference   \\
				\hline
				1       &GM(1,1)     &$\frac{d x^{(1)}(t)}{d t}+a x^{(1)}(t)=b$    & \cite{lin2004historical}   \\
				2       &NGM(1,1,$k$,$c$)       &$x^{(0)}(k)=a x^{(0)}(k-1)+b$                 & \cite{chen2014foundation}  \\
				3       &GM(1,1,${{t}^{\alpha}}$)        &$\frac{d x^{(1)}(t)}{d t}+a x^{(1)}(t)=b t^{\alpha}+c, r>0, \alpha>0$                 & \cite{qian2012grey}  \\
				4       &GMP(1,1,$N$)         &$\frac{d x^{(1)}(t)}{d t}+a x^{(1)}(t)=\beta_{0}+\beta_{1} t+\beta_{2} t^{2}+\cdots+\beta_{\mu} t^{n}$     & \cite{dang2017grey}  \\
				5       &GVM(1,1)      & $\sum_{i=k}^{n} x^{(0)}(i)=a \sum_{i=k}^{n} x^{(0)}(i-1)+(n-k+1) b$ & \cite{zhang2012improved} \\
				6       & NGBM(1,1)  & $\frac{\mathrm{d} x^{(1)}(t)}{\mathrm{d} t}+a x^{(1)}(t)=b\left(x^{(1)}(t)\right)^{\gamma}$      & \cite{chen2008forecasting} \\
				7       &GBM(1,1)    & $\frac{d x^{(1)}(t)}{d t}=a\left(M-x^{(1)}(t)\right)+b x^{(1)}(t)\left(1-\frac{x^{(1)}(t)}{M}\right)$             & \cite{bass1969new}\\
				\hline
				\label{tab1}
	\end{tabular} }}
\end{table*}

\subsubsection*{4. GMP(1,1,$N$)}
In 2017, Luo and Wei \cite{dang2017grey} proposed a grey model with a polynomial term called GMP(1,1,$N$) where the grey action term is a time polynomial function. The GM(1,1) model, the NGM(1,1,$k$) model, and the GM(1,1,${{t}^{\alpha}}$) model have been shown to be special cases of the GMP(1,1,$N$) model. The differential form of the GMP (1,1,$N$) model obtained as,
\begin{align}
\frac{d x^{(1)}(t)}{d t}+a x^{(1)}(t)=\beta_{0}+\beta_{1} t+\beta_{2} t^{2}+\cdots+\beta_{\mu} t^{n},
\end{align}
Whereas the time response function for the GMP (1,1,$N$) is,
\begin{align}
x^{(1)}(k)=\left(x^{(0)}(1)-\sum_{r=0}^{k} r_{i}\right) e^{-a(k-1)}+\sum_{i=0}^{N}\left(r^{i} k^{i}\right),
\end{align}
Finally, the restored value for the GMP (1,1,$N$) is,
\begin{align}
x^{(0)}(k)=x^{(1)}(k)-x^{(1)}(k-1).
\end{align}

\subsubsection*{5. GVM(1,1)}
The Grey Verhulst model GVM(1,1) is suitable for forecasting the frequency of sequences that have a single apex or whose development delayed \cite{wang2009unbiased}. Suppose that, there is a positive sequence of data $X^{(0)}=\left\{x^{(0)}(1), x^{(0)}(2), \ldots, x^{(0)}(n)\right\}$, and $\quad X^{(1)}$ is accumulated generating operation (AGO) of $X^{(0)}$, written as $X^{(1)}=\left\{x^{(1)}(1), x^{(1)}(2), \ldots, x^{(1)}(n)\right\}$, where, $x^{(1)}(k)=\sum_{i=1}^{K} x^{(0)}(i)$, $k=1,2, \ldots,n$. $Z^{(1)}=\left\{z^{(1)}(1), z^{(1)}(2), \ldots, z^{(1)}(n)\right\}$ is the mean sequence of $x^{(1)}(k)$, while, $z^{(1)}(k)=\frac{1}{2}\left(x^{(1)}(k)+x^{(1)}(k-1)\right)$, $k=2,3, \dots, n$ \cite{zhang2019application}.\\
\begin{align}
y^{(1)}(k+1)=\beta_{0}+\beta_{1} k+\beta_{2} y^{(1)}(k),
\label{eq:18}
\end{align}
Equation (18) is an optimized discrete Verhulst model. Whereas the differential equation of the conventional grey verhulst model \cite{zhang2012improved} is given below:
\begin{align}
\frac{d\left(x^{(1)}\right)}{d t}+a x^{(1)}=b\left(x^{(1)}\right)^{2},
\end{align}
Whereas the time response function for Grey Verhulst model is,
\begin{align}
x^{(1)}(k+1)=\frac{a x^{(1)}(0)}{b x^{(1)}(0)+\left(a-b x^{(1)}(0)\right) E X P(a k)},
\end{align}
while the restored value of grey verhulst model is,
\begin{align}
\hat{x}^{(0)}(k)=\hat{x}^{(1)}(k)-\hat{x}^{(1)}(k-1) ,	 k=2,3, \cdots, n.
\end{align}

\subsubsection*{6. NGBM(1,1)}
The nonlinear grey Bernoulli model NGBM(1,1) is a branch of the Grey Verhulst model and GM(1,1) \cite{chen2008application,2020B}. The key advantage of NGBM(1,1) is the power exponents of the model will effectively represent the non-linear features of the true structure and evaluate the form of the model in a versatile manner. Therefore, the framework can not only forecast sequences that increase or decrease monotonically but can also respond favorably to nonlinear processes with small sample sizes \cite{xie2020novel,chen2008forecasting}. The parameters of the NGBM(1,1) are also calculated by the least square estimation approach known as the Levenberge Marquardt (LM) optimization theory, and then use the modified model to allow analytical comparisons with the regression models, the traditional GM(1,1) and the Grey Verhulst models, which suggests that NGBM(1,1) shows the desirable prediction accuracy \cite{shaikh2017forecasting,ma2017novel}. Therefore, it is empirically proved that the performance of the NGBM(1,1) is higher than that of the GM(1,1) and Grey Verhulst model \cite{pei2018nls}.
The differential equation of the NGBM(1,1) model is given by \cite{chen2008forecasting},
\begin{align}
\frac{\mathrm{d} x^{(1)}(t)}{\mathrm{d} t}+a x^{(1)}(t)=b\left(x^{(1)}(t)\right)^{\gamma},
\end{align}
Furthermore, the time response function of the NGBM(1,1) model is obtained as,
\begin{align}
x^{(r)}(t)=\left[\left(\left(x^{(r)}(1)\right)^{1-\gamma}-\frac{b}{a}\right) e^{-a(1-\gamma)(t-1)}+\frac{b}{a}\right]^{\frac{1}{1-\gamma}},
\end{align}
while the prediction value of NGBM(1,1) is,
\begin{align}
\begin{array}{r}{x^{(r)}(k)=\left[\left(\left(x^{(r)}(1)\right)^{1-\gamma}-\frac{b}{a}\right) e^{-a(1-\gamma)(k-1)}+\frac{b}{a}\right]^{\frac{1}{1-\gamma}}} \\ {k=2,3, \ldots, n}.\end{array}
\end{align}

\subsubsection*{7. GBM(1,1)}
In 1969, Bass et al. proposed once the purchasing model of durable goods by market research on the prevalence of 11 types of durable goods, abbreviated as a Grey Bass model GBM(1,1) \cite{bass1969new}. Due to the simple form and the clear economic sense of the parameters, the Bass model is commonly used in new product forecasts \cite{scaglione2015diffusion,lee2014pre},  technological diffusion \cite{sood2012predicting}, and business model diffusion \cite{marshall2013forecasting,seol2012demand,wang2013study}.
Moreover, the mathematical form of the grey bass model is,
\begin{align}
\frac{d x^{(1)}(t)}{d t}=a\left(M-x^{(1)}(t)\right)+b x^{(1)}(t)\left(1-\frac{x^{(1)}(t)}{M}\right),
\end{align}
while the time response function of the grey bass model is,
\begin{align}
\hat{x}^{(1)}(k)=M\left[\frac{1-e^{-(a+b) k}}{1+\frac{b}{a} e^{-(a+b) k}}\right], k=1,2, \cdots, n.
\end{align}
At last, the restored value of the grey bass model is,
\begin{align}
\hat{x}^{(0)}(k+1)=\hat{x}^{(1)}(k+1)-\hat{x}^{(1)}(k), k=1,2, \cdots, n.
\end{align}

\subsubsection{\textbf{Multivariate Grey Models}}
The multivariate grey forecasting model is represented by GM(1,$n$). This model consists of system-specific sequences (or dependent variable sequences) and ($n$-1) related sequences of variables (or independent variable sequences). The modeling method takes complete account of the effect of relevant variables on system transition and is a standard causal forecasting model. The GM(1,$n$) model has some similarities to the multi-regression model, yet it is fundamentally distinct. The former is focused on grey theory, while the latter is centered on probability statistics. The multivariate gray forecasting model is the limitation of the single-fold framework and the limited simulation potential of single-variable models. This model is mainly used as a tool for determining the similarities between system features sequences and associated factor sequences \cite{jie2011grey,ker2011using,wang2014gm,chen2011applying,hao2013piecewise}. The detailed differential equation of multivariate grey forecasting models have been summarized in Table \ref{tab2}

\subsubsection*{1. GM(1,$n$)}
GM(1,$n$) demonstrated that the approximate whitening time response function of GM(1,$n$) could often contribute to an inappropriate experimental error \cite{tien2012research,wang2017forecasting}. Furthermore, the solution of the whitening equation $\frac{d x_{1}^{(1)}}{d t}+a x_{1}^{(1)}=\sum_{i=2}^{N} b_{i} x_{i}^{(1)}$ is defined by,
 \begin{align}
\begin{aligned} &x_{1}^{(1)}(t)=e^{-at}\\
&\hspace{5mm}\left[x_{1}^{(1)}(0)-t \sum_{i=2}^{N} b_{i} x_{i}^{(1)}(0)+\sum_{i=2}^{N} \int b_{i} x_{i}^{(1)}(t) e^{a t} d t\right],
 \end{aligned}
 \end{align}
When $X_{i}^{(1)}(i=1,2, \ldots, N)$ changes marginally, $\sum_{i=2}^{N} b_{i} x_{i}^{(1)}(k)$ is shown as a grey constant. While the estimated time-response sequence of the GM(1,$n$) model is obtained as,
\begin{align}
\hat{x}_{1}^{(1)}(k+1)=\left[x_{1}^{(1)}(0)-\frac{1}{a} \sum_{i=2}^{N} b_{i} x_{i}^{(1)}(k+1)\right] e^{-a k}\nonumber\\
+\frac{1}{a} \sum_{i=2}^{N} b_{i} x_{i}^{(1)}(k+1),
\end{align}
where $x_{1}^{(1)}(0)$ is assumed to be $x_{1}^{(0)}(1)$, which is the initial value of the GM(1,$n$) model. Restoration of the inverse accumulation of GM(1,$n$) is obtained by,
\begin{align}
\hat{x}_{1}^{(0)}(k+1)=\alpha^{(1)} \hat{x}_{1}^{(1)}(k+1)=\hat{x}_{1}^{(1)}(k+1)-\hat{x}_{1}^{(1)}(k).
\end{align}

\subsubsection*{2. GMC(1,$n$)}
To compare GMC(1,$n$) with other prediction models, the GMC(1,$n$) model requires a greater proportion of the forecast core data and is far more significant than the other forecast source data. The accuracy of the indirect estimation and forecast of the GMC(1,$n$) model can then be presumed. This framework is used to test indirect measurements which cannot be sufficiently accurate for the needs of the analyzer. Since the analyzer can find it appropriate to search for external guidance factors which can improve the accuracy to a certain level \cite{tien2005indirect}. The authors suggested that the GMC(1,$n$) model has the following linear differential equation,
\begin{align}
x_{1}^{(0)}(r p+t)+b_{1} z_{1}^{(1)}(r p+t)=\sum_{i=2}^{n} b_{i} z_{i}^{(1)}(t)+u,
\end{align}
while the time response function of GMC(1,$n$) can be obtained by,
\begin{align}
&\hat{x}_{1}^{(1)}(r p+t)=x_{1}^{(0)}(r p+1) e^{-b_{1}(t-1)}\nonumber\\
&\hspace{3.5cm}+\int_{1}^{t} e^{-b_{1}(t-\tau)} f(\tau) d t,
\end{align}
and the predicted value of GMC(1,$n$) is given as,
\begin{align}
\hat{x}_{1}^{(0)}(r p+t)=\hat{x}_{1}^{(1)}(r p+t)-\hat{x}_{1}^{(1)}(r p+t-1).
\end{align}

\subsubsection*{3. NGMC(1,$n$)}
The NGMC(1,$n$) model effectively describes and achieves satisfactory predictability as compared to the traditional GMC(1,$n$), and $n$-1 power exponents of the expected variables. The $\beta_{2}, \beta_{3}, \dots, \beta_{n}$ was introduced in NGMC(1,$n$) model to represent the impact of these upon nonlinear system behaviors and interactions. Unknown parameters are calculated by a computer program that calculates the minimum average relative percentage error of the forecasting model. This enhances the adaptability of the NGMC(1,$n$) model to the initial data and consequently strengthens the accuracy of the prediction \cite{wang2014nonlinear}. Nonetheless, the current fractional multivariate grey model with convolution integral is known as the following linear differential equation,
\begin{align}
\frac{d x_{1}^{(r)}(t)}{d t}+b_{1} x_{1}^{(r)}(t)=\sum_{i=2}^{n} b_{i} x_{i}^{(r)}(t)+u,
\end{align}
whereas the continuous time response function can be obtained as,
\begin{align}
\hat{x}_{1}^{(1)}(t)=x_{1}^{(0)}(1) e^{-b_{1}(t-1)}+\int_{1}^{t} e^{-b_{1}(t-\tau)} f(\tau) d t,
\end{align}
Finally, the predicted value of NGMC(1,$n$) can be obtained using the inverse fractional-order accumulation as,
\begin{align}
&\hat{x}_{1}^{(0)}(k)=\left(\hat{x}^{(r)}(k)\right)^{(-r)}\nonumber\\
&\hspace{18mm}=\sum_{i=1}^{k}\left(\begin{array}{c}{k-i-r-1} \\ {k-i}\end{array}\right) \hat{X}^{(r)}(i).
\end{align}

\begin{table*}[!htbp]
	\centering
	\caption{The definition of the comparative Multivariate grey models used.}
	\renewcommand{\baselinestretch}{2.5}
	{\footnotesize\centerline{\tabcolsep=6pt
			\begin{tabular}{cccccccccccccc}
				\hline
				No.     &    Model Name           &   Differential Equation    &  Reference   \\
				\hline
				1       &GM(1,$n$)     &$[x_{1}^{(1)}(0)-t \sum_{i=2}^{N} b_{i} x_{i}^{(1)}(0)+\sum_{i=2}^{N} \int b_{i} x_{i}^{(1)}(t) e^{a t} d t$    & \cite{tien2012research}   \\
				2       &GMC(1,$n$)       &$x_{1}^{(0)}(r p+t)+b_{1} z_{1}^{(1)}(r p+t)=\sum_{i=2}^{n} b_{i} z_{i}^{(1)}(t)+u$                 & \cite{tien2005indirect}    \\
				3       &NGMC(1,$n$)        &$\frac{d X_{1}^{(r)}(t)}{d t}+b_{1} x_{1}^{(r)}(t)=\sum_{i=2}^{n} b_{i} x_{i}^{(r)}(t)+u$                 & \cite{wang2014nonlinear}\\
				4       &RDGM(1,$n$)         &$x_{1}^{(1)}(r p+k+1)=\beta_{1} x_{1}^{(1)}(r p+k)+\sum_{i=2}^{n} \beta_{i} z_{i}^{(1)}(k+1)+\mu$         & \cite{ma2016research}  \\
				5       &OGM(1,$n$)      & $\hat{x}_{1}^{(0)}(k)=\sum_{i=2}^{N} b_{i} \hat{x}_{i}^{(1)}(k)-a z_{1}^{(1)}(k)+h_{1}(k-1)+h_{2}$ & \cite{zeng2016development} \\
				6       & NGM(1,$n$)  & $\frac{d x_{1}^{(1)}(t)}{d t}+a x_{1}^{(1)}(t)=\sum_{i=2}^{N} b_{i}\left(x_{i}^{(1)}(t)\right)^{\gamma_{i}}$    & \cite{wang2017forecasting} \\
				7       &KGM(1,$n$)   &$x_{1}^{(0)}(k)+a z_{1}^{(1)}(k)=\phi(k)+u$ & \cite{ma2018kernel}\\
				\hline
				\label{tab2}
	\end{tabular} }}
\end{table*}

\subsubsection*{4. RDGM(1,$n$)}
In 2016, Xin Ma et al. \cite{ma2016research} presented a novel multivariate grey prediction model represented by RDGM(1,$n$). In which he proposed the modeling  procedures in detail. The statistical analysis was introduced to demonstrate the interaction and gap between the traditional GMC(1,$n$) and RDGM(1,$n$) as well as the numerical examples were used to validate the output of RDGM(1,$n$) compared with GMC(1,$n$). The shortcomings of RDGM(1,$n$) were highlighted by the author. The RDGM(1,$n$) was basically a linear model. Therefore, it can not be feasible to explain the cause and effect of nonlinear processes for which inputs and outputs are non-linearly linked. The basic form of RDGM(1,$n$) is obtained as,
\begin{align}
&x_{1}^{(1)}(r p+k+1)=\beta_{1} x_{1}^{(1)}(r p+k)\nonumber\\
&\hspace{3cm}+\sum_{i=2}^{n} \beta_{i} z_{i}^{(1)}(k+1)+\mu,
\end{align}
where the time response function of  RDGM(1,$n$) is then specified as,
\begin{align}
&x_{1}^{(1)}(r p+k+1)=\frac{1-0.5 b_{1}}{1+0.5 b_{1}} x_{1}^{(1)}(r p+k)\nonumber\\
&\hspace{12mm}+\sum_{i=2}^{n} \frac{b_{i}}{1+0.5 b_{1}} z_{i}^{(1)}(k+1)+\frac{u}{1+0.5 b_{1}},
\end{align}
while the predicted series of RDGM(1,$n$) can be obtained as,
\begin{align}
\hat{x}_{1}^{(0)}(r p+k+1)=\hat{x}_{1}^{(1)}(r p+k+1)-\hat{x}_{1}^{(1)}(r p+k).
\end{align}

\subsubsection*{5. OGM(1,$n$)}
In 2016, Zeng et al. \cite{zeng2016development} were presented a new multivariable grey optimizing model namely called OGM(1,$n$). The main purpose of this model is to solve the mechanism defect, parameter, and structural defect found in the conventional GM(1,$n$) model. Theoretically, it has been shown that the GM(0,$n$), DGM(1,1) and NDGM(1,1) models are all similar variants of the OGM(1,$n$) model with varying parameter values. The authors proposed that the differential equation of OGM(1,$n$) can be obtained as,
\begin{align}
\hat{x}_{1}^{(0)}(k)=\sum_{i=2}^{N} b_{i} \hat{x}_{i}^{(1)}(k)-a z_{1}^{(1)}(k)+h_{1}(k-1)+h_{2},
\end{align}
where the time-response expression is given by,
\begin{align}
&\hat{x}_{1}^{(1)}(k)= \sum_{t=1}^{k-1}\left[\mu_{1} \sum_{i=2}^{N} \mu_{2}^{t-1} b_{i} x_{i}^{(1)}(k-t+1)\right]\nonumber\\
&\hspace{18mm}+\mu_{2}^{k-1} \hat{x}_{1}^{(1)}(1)+\sum_{j=0}^{k-2} \mu_{2}^{j}\left[(k-j) \mu_{3}+\mu_{4}\right], \nonumber\\
&\hspace{5.5cm}k=2,3, \cdots
\end{align}
and, its restoration expression of inverse accumulation is obtained as,
\begin{align}
\hat{x}_{1}^{(0)}(k)=\sum_{i=2}^{N} b_{i} \hat{x}_{i}^{(1)}(k)-a z_{1}^{(1)}(k)+h_{1}(k-1)+h_{2}.
\end{align}

\subsubsection*{6. NGM(1,$n$)}
To characterize the NGM(1,$n$) and its transformed model, a more precise explanation is important about nonlinear interaction between the variables relative to GM(1,$n$) and to enhance the accuracy of the forecasting of the individual data sequences \cite{wang2017forecasting}. Furthermore, the authors presented the solution of the whitening equation as,
\begin{align}
\frac{d x_{1}^{(1)}(t)}{d t}+a x_{1}^{(1)}(t)=\sum_{i=2}^{N} b_{i}\left(x_{i}^{(1)}(t)\right)^{\gamma_{i}},
\end{align}
whereas the time response sequence of the NGM(1,$n$) model is then specified as,
\begin{align}
\widehat{x}_{1}^{(1)}(k+1)=\left[x_{1}^{(1)}(1)-\frac{1}{a} \sum_{i=2}^{N} b_{i}\left(x_{i}^{(1)}(k+1)\right)^{\gamma_{i}}\right] e^{-a k} \nonumber\\
+\frac{1}{a} \sum_{i=2}^{N} b_{i}\left(x_{i}^{(1)}(k+1)\right)^{\gamma_{i}},
\end{align}
The restoration of inverse accumulation is obtained as,
\begin{align}
\widehat{x}_{1}^{(0)}(k+1)=\alpha^{(1)} \widehat{x}_{1}^{(1)}(k+1)=\widehat{x}_{1}^{(1)}(k+1)-\widehat{x}_{1}^{(1)}(k).
\end{align}

\subsubsection*{7. KGM(1,$n$)}
In 2018, Xin Ma et al. \cite{ma2018kernel} introduced a novel nonlinear multivariate grey model which is based on the kernel method, namely kernel-based GM(1,$n$). Which is also known as KGM(1,$n$). The key feature of this model is to include an unknown component of the input series, which can be calculated using the kernel function, and then the KGM(1,$n$) model is used to explain the nonlinear relationship between the input and output sequence. The differential equation of KGM(1,$n$) is,
\begin{align}
x_{1}^{(0)}(k)+a z_{1}^{(1)}(k)=\phi(k)+u,
\end{align}
while the time response function of KGM(1,$n$) can be written as,
\begin{align}
\hat{x}_{1}^{(1)}(k)=\alpha^{k-1} x_{1}^{(0)}(1)+\sum_{\tau=2}^{k}(\psi(\tau)+\mu) \cdot \alpha^{k-\tau},
\end{align}
and the restored values can be obtained as,
\begin{align}
\hat{x}_{1}^{(0)}(k)=\hat{x}_{1}^{(1)}(k)-\hat{x}_{1}^{(1)}(k-1).
\end{align}

The main structure is also described as shown in the grey model. Besides the free parameters $\alpha$ and $\boldsymbol{\theta}$, the grey model components are not specified. Essentially, the arrangement of $f(\boldsymbol{\theta}; t)$ or $f(\boldsymbol{\theta}; k)$ also plays a key role in enhancing grey models. It is well-known that the non-homogeneous grey NGM model often performs marginally better than the GM(1,1) and related results can be widely found in current studies. Furthermore, the optimum formulation of the $f(\cdot)$ function cannot be always consider in real-world applications. This question contributes to give inspiration to the researchers to apply Machine Learning over Grey Models.

\section{\textbf{Grey Machine Learning (GML)}}
\label{sec:GML}
Grey Machine Learning (GML) is a multivariate model based on kernel for time series forecasting, suggested by Xin Ma et al. in 2018 \cite{Ma2018ABI}. The main purpose is to investigate and highlight the capability of GML on time series forecasting. GML framework is a kernel-based multivariate model hence it should be used to compensate between the models and data. The detailed timeline of the GML framework have been discussed in Figure \ref{fig7}.

\subsection{Overview of GML}
This brief overview is based on the question of, why the researchers need a reliable framework to train machines for time series datasets. In 2015, Brenden Lake et al. presented a machine to understand Mongolian and passed the Turing test \cite{lake2015human}. Such research has shown that the computer can function as humans in handwriting tasks. Another breakthrough is Google DeepMind's AlphaGo. In 2016, David Silver et al. published AlphaGo's first paper, stating that the machine could beat the Go game practitioners for the very first time, and this article was published after AlphaGo defeated the World Champion Lee Sedol. The success of AlphaGo has also demonstrated that the machine can not only be like humans, but can also be smarter than humans, and the output of this research gives AI enormous confidence \cite{long2019novel,silver2016mastering}.

Defining the role of AI is about creating machines with intelligence, like humans or animals. Mathematical models and algorithms are important for achieving that goal. In the research mentioned above, Brenden Lake et al. used the Bayesian System for handwriting activities, and AlphaGo is built on the neural networks and Monte Carto tree search. Nevertheless, such studies also have other challenges, such as they required a large amount of training data to achieve the best models. It is also really interesting to note that Google has built another Go game machine named AlphaGo Zero \cite{silver2017mastering}. This revolutionary machine trained without any chess manual, only need to teach by using the rules of the Go game. Thus, AlphaGo Zero won the championship successfully and beat AlphaGo \cite{silver2016mastering}.

\Figure[t!][width=5.0in,height=5.2in,clip,keepaspectratio]{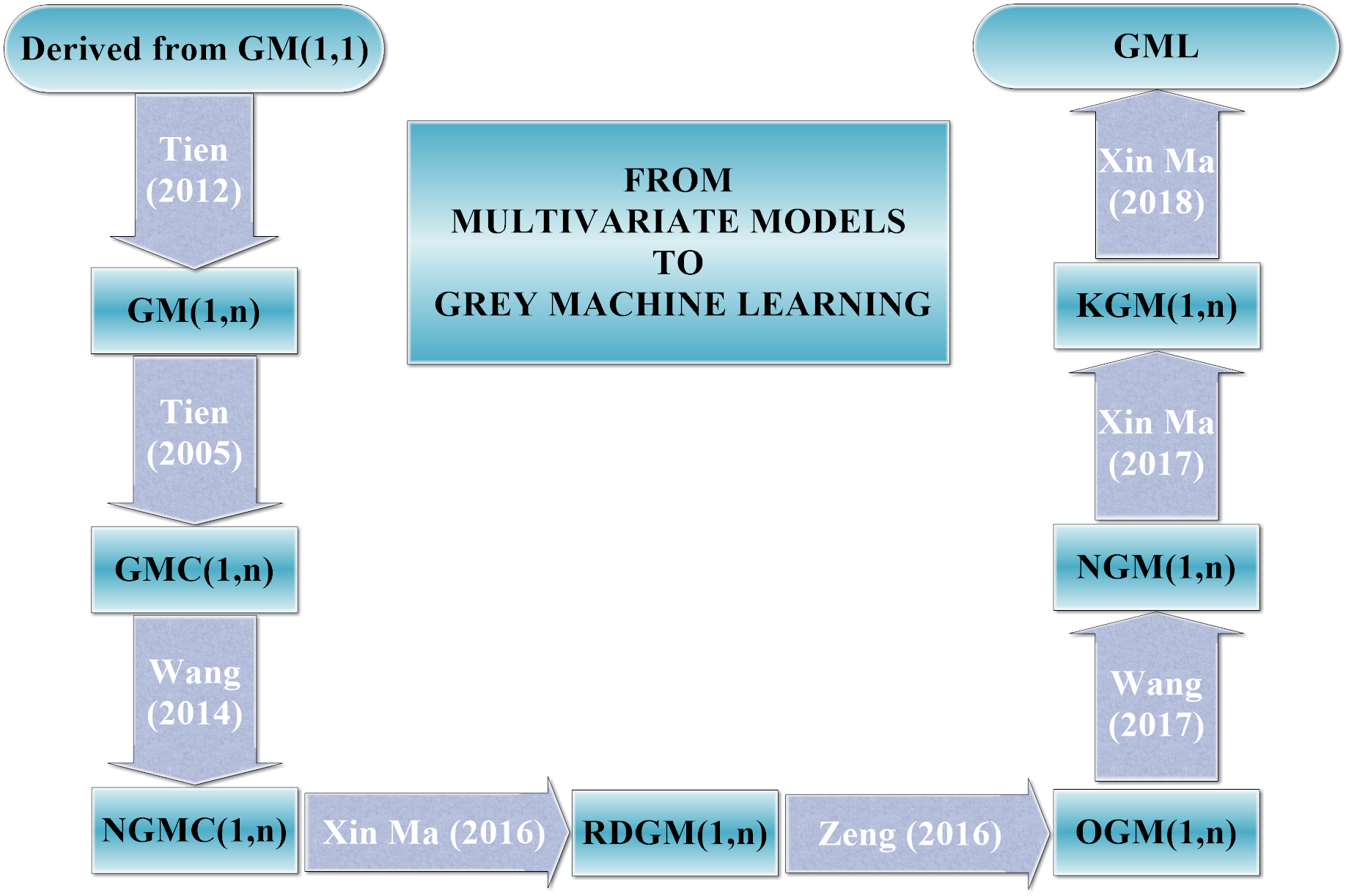}
{Timeline of the Grey Machine Learning.\label{fig7}}

The above studies demonstrated that, it is attainable to train machines utilizing very small time series datasets. This means that Big Data may not be the only way to make AI efficient. Also, researchers have often observed very small data cases in the process of human learning. For starters, a little kid learns the basic techniques of the Go game with a certain lessons. Scientifically, the main methodology used to improve AlphaGo is Deep Learning and Improving training, both of these are focused on the concept of getting the best of human intelligence. Moreover, the variants in the Go game are computationally limitless for computers, although contrary to all instances, the chess guides provided for AlphaGo are restricted. It is thus difficult to train a smart AlphaGo without effective learning methods focused on small samples. Grey system theory is primarily designed for limited datasets and also appears to have a deterministic structure with free parameters to be determined by samples \cite{2020C}.

This concept is derived from system science, the deterministic configuration of grey models is an established part of the system, with free parameters as an undefined component. Such kind of phenomenon is quite limiting, as it is really hard to discover the complexity of systems in real-world applications. In most situations, only a small part of the devices can be identified in a limited time or at the least cost. It is obvious to ask the question-how can structures be modelled given that they are partly understood?

Based on this question, Xin Ma carried out a range of experiments to merge the principles of grey modeling with the techniques of Machine Learning. The main issues discussed in these studies are structures with a defined complex configuration and an undefined nonlinear interaction between the device state or the output and input series. For instance, previous results in real-world applications also suggested that models built on this concept are significantly more efficient than traditional grey models \cite{Ma2018ABI,Cerqueira2020EvaluatingTS}. A brief flowchart of GML framework have been discussed in Figure \ref{fig8}.

\subsection{\textbf{General Formulation of GML}}
The general structure of the GML model is classified into the following equations proposed by \cite{Ma2018ABI}:
\begin{itemize}
	\item \textbf{The continuous Structure:}
\end{itemize}
\begin{align}
\frac{d x_{1}^{(1)}(t)}{d t}+a x_{1}^{(1)}(t)=\phi(t)+u
\label{eq:48}
\end{align}
\begin{itemize}
	\item \textbf{The discrete structure:}
\end{itemize}
\begin{align}
x_{1}^{(1)}(k+1)=\alpha x_{1}^{(1)}(k)+\phi(k)+\mu
\label{eq:49}
\end{align}
While the element $\phi(\cdot)$ is completely unknown, and this is the main difference between GML and conventional grey models.

\subsection{\textbf{An Estimate of uncertain function}} 
There are several phases to approximate the uncertain function $\phi(\cdot)$ in the general formulation of the GML model\cite{ma2016research,Ma2018ABI}, which are given below:
\subsubsection{\textbf{Linear representation in the higher dimensional space function}}
Based on the Theorem of Weierstrass \cite{Bialas99thetheorem}, every continuous function at a closed and bound time can be uniformly approximated by polynomials to some degree of accuracy. Furthermore, these are the following forms of the function as given below:
\begin{align}
\phi(t)=w_{n} x^{n}(t)+w_{n-1} x^{n-1}(t)+\ldots+w_{1} x^{1}(t)+w_{0}
\label{eq:50}
\end{align}
Lets the mapping be $\varphi= R \rightarrow \mathscr{F}$, where $\mathscr{F}=\left\{\left[x^{n}(t), x^{n-1}(t), \ldots, x^{1}(t), 1\right]^{T}\right\}$,  and the weights are $\omega=\left[w_{n}, w_{n-1}, \dots, w_{1}, w_{0}\right]^{T}$. The nonlinear equation can then be written in linear form in the higher dimensional space $\mathscr{F}$ as,
\begin{align}
\phi(t)=\omega^{T} \varphi(x(t)).
\label{eq:51}
\end{align}
Remember that this definition still holds only $x(t)$ as a function, and the range of the $\mathscr{F}$ element can be infinite.

\subsubsection{\textbf{Nonparametric Estimation}}
In the above example, the estimation of arbitrary nonlinear equations can be transformed into a linear problem. With the specified samples $\{(x(k), y(k)) \mid k=1,2,3, \ldots, N\}$, which can be estimation from the following formulation \cite{Ma2018ABI},
\begin{align}
y(t)=\omega^{T} \varphi(x(t))+b.
\label{eq:52}
\end{align}
The issue of regularization is also in the shape of ridge regression is defined as,
\begin{align}
 &\min J(\omega, e) =\frac{\|\omega\|^{2}}{2}+\frac{C}{2} \sum_{k=1}^{m} e_{k}^{2} \nonumber\\
&\hspace{18mm}\text { s.t. }  e_{k}=y(k)-\omega^{T} \varphi(x(k))-b 
\end{align}
Where the value $\|\cdot\|$ is the 2-norm. The key distinction between the problem of regularization (54) and the commonly adopted least-square solution is that the $\|\omega\|^{2}$ should always be small. The nonlinear mapping $\varphi$ is not unique. For example, when the nonlinear equation $\phi(t)$ becomes discontinuous for a higher value, it can be generalized for the Taylor Power Series. Therefore, there are various variants where $\phi(t)$ becomes applied at specific stages. Mathematically, a brief analysis is required to guarantee that the response is special. Whereas, with the non-deterministic mathematical functions, the problem of regularization is often classified in nonparametric estimation formulations. On the other hand, researchers also want the forecast to be smooth enough to make generalization easier or more accurate. The term $C$ should be used for this reason. The question of regularization (54) is restricted a quadric programming. Therefore, the Lagrangian should need to describe as a first step.
\begin{align}
&L:=\frac{\|\omega\|^{2}}{2}+\frac{C}{2} \sum_{k=1}^{r} e_{k}^{2}\nonumber\\
&\hspace{1cm}+\sum_{k=1}^{m} \lambda_{k}\left[y(k)-\omega^{T} \phi(x(k))-b-e_{k}\right].
\end{align}
The solution to the problem of regularization (54) can be easily achieved by implementing the Karush-Kuhn-Tucker (KKT) conditions as follows:
\begin{align}
\cr\left\{\begin{array}{l}{\frac{\partial L}{\partial w}=0 \Rightarrow \omega=\sum_{k=1}^{m} \lambda_{k} \varphi(x(k))} \\ 
\cr{\frac{\partial L}{\partial b}=0 \Rightarrow \sum_{k=1}^{m} \lambda_{k}=0} \\ 
\cr{\frac{\partial L}{\partial \lambda_{j}}=0 \Rightarrow e_{k}=\lambda_{k} / C} \\ 
\cr{\frac{\partial L}{\partial \lambda_{j}}=0 \Rightarrow y(k)-\omega^{T} \varphi(\chi(k))-b=e_{k}.}\end{array}\right.
\end{align}

Equation (56) indicates that $\omega$ can be determined using the values of the Lagrangian multipliers $\lambda_{k}$ and $\varphi(x(k))$. As a consequence, the computational representation of the nonlinear function can be obtained as,

\begin{align}
\omega^{T} \varphi(x(t))=\sum_{k=1}^{m} \lambda_{k} \varphi^{T}(x(k)) \varphi(x(t)).
\label{eq:56}
\end{align}

Although, the author \cite{Ma2018ABI} has not yet provided the deterministic form of the nonlinear mapping of $\varphi(\cdot)$. Although, it is only need to calculate the inner product $<\varphi(x(k)), \varphi(x(t))>=\varphi^{T}(x(k)) \varphi(x(t))$ in the $\mathscr{F}$ function. Therefore, if the Mercer conditions under which realize that the author has the following expansion for any symmetric positive specified function $K(\cdot,\cdot)$, respectively.

\begin{align}
K(x(k), x(t))=\sum_{n=1}^{\infty} \alpha_{n} \psi_{n}(x(k)) \psi_{n}(x(t)),
\label{eq:57}
\end{align}

while $\left\{\psi_{n}(x(t))\right\} \begin{array}{l}{n=\infty} \\ {n=1}\end{array}$ is the orthogonal basis, and $\alpha_{n}$ is the positive eigenvalues. Such features are also referred to as core functions or kernels.
Therefore, the nonlinear mapping corresponding to the given kernel can be easily defined as,
\begin{align}
\varphi(\cdot)=\left[\sqrt{\alpha_{1}} \psi_{1}(\cdot), \sqrt{\alpha_{2}} \psi_{2}(\cdot), \ldots, \sqrt{\alpha_{n}} \psi_{n}(\cdot), \ldots\right]^{T}.
\end{align}
Clearly, the inner product of nonlinear projection can be described as a kernel,
\begin{align}
 &K(x(k), x(t)) =\sum_{n=1}^{\infty} \alpha_{n} \psi_{n}(x(k)) \psi_{n}(x(t)) \nonumber\\ 
 &\hspace{7mm}=<\left[\sqrt{\alpha_{1}} \psi_{1}(k), \sqrt{\alpha_{2}} \psi_{2}(k), \ldots, \sqrt{\alpha_{n}} \psi_{n}(k), \ldots\right]^{T},\nonumber\\
 &\hspace{10mm}\left[\sqrt{\alpha_{1}} \psi_{1}(t), \sqrt{\alpha_{2}} \psi_{2}(t), \ldots, \sqrt{\alpha_{n}} \psi_{n}(t), \ldots\right]^{T}>\nonumber\\ 
 &\hspace{4cm}=\varphi^{T}(x(k)) \varphi(x(t)). 
\end{align}
Therefore, through substituting (57), (62) into (56) with the exclusion of $e_{k}$, the solution of the regularized problem is expressed by the following linear framework,
\begin{align}
\left(\begin{array}{ccc}
0 & \vdots & 1^T_m\\
\cdots & \cdots & \cdots\\
1_{m}& \vdots & \Omega+I_{m} / C
\end{array}\right)\left(\begin{array}{c}
b \\
\cdots \\
\lambda
\end{array}\right)=\left(\begin{array}{c}
0 \\
\cdots \\
Y
\end{array}\right)
\label{eq:60}
\end{align}
where,
\begin{align}
\mathbf{1}_{m}=[1,1, \ldots, 1]_{m}^{T}, \nonumber\\ 
\Omega=[K(x(i), x(j))]_{m \times m}, \nonumber\\ 
\boldsymbol{\lambda}=\left[\lambda_{1}, \lambda_{2}, \lambda_{3}, \ldots, \lambda_{m}\right]^{T}, \nonumber\\ 
Y=[y(1), y(2), \ldots, y(m)]^{T},\nonumber
\end{align}\\
and $I_{m}$ is an identity matrix.\\
It can be shown that the solution of the regularized problem (54) can be transformed to a linear model (61), which can be quite easy to solve with a given kernel, and the coefficients are $\lambda$ and $b$.

At last point, the statistical formulation of the estimation needs to be provided (53). Under (57) and (62), the estimation of (53) can be obtained as,
\begin{align}
\begin{aligned} \hat{y}(t) &=\omega^{T} \varphi(x(t))+b \\ &=\sum_{k=1}^{m} \lambda_{k} \varphi^{T}(x(k)) \varphi(x(t))+b \\ &=\sum_{k=1}^{m} \lambda_{k} K(x(k), x(t))+b \end{aligned}
\label{eq:61}
\end{align}

  \Figure[t!][width=4.5in,height=5in,clip,keepaspectratio]{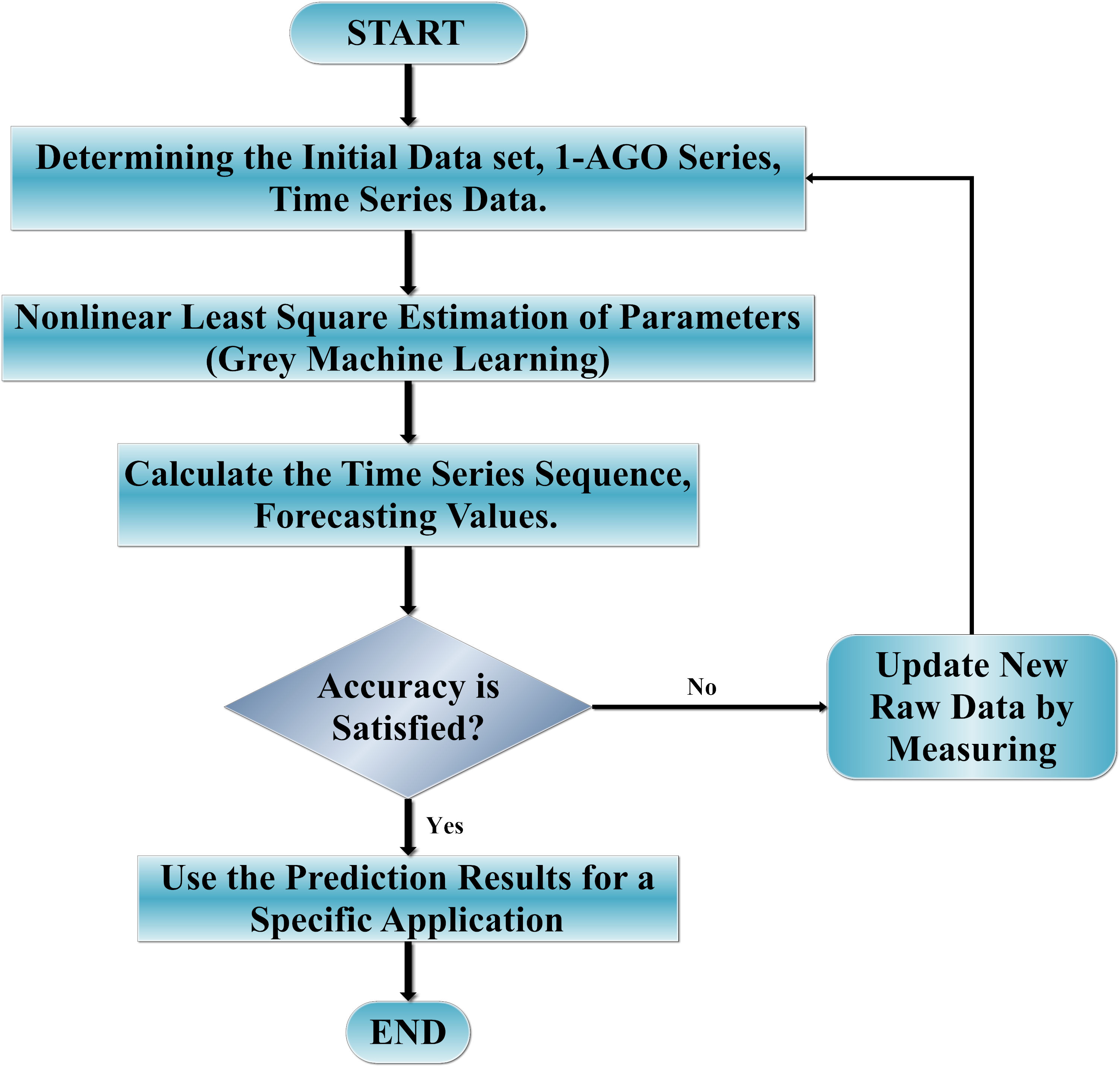}
{Flowchart of the Grey Machine Learning.\label{fig8}} 

\subsubsection{\textbf{Semi-parametric Estimation}}
With a nonparametric estimation, it is easy to deduce the semi-parametric variant of which the goal is to make an approximation that uses the following terminology,
\begin{align}
y(t)=\omega_{1}^{T} z(t)+\omega_{2}^{T} \varphi(x(t))+b,
\label{eq:62}
\end{align}
while $z(t)$ is the variable in $R$ or the parameter in linear space $R^{N_{1}}$. This form can be converted to a nonparametric formulation by representing a new nonlinear mapping of $\varphi^{\prime}: R^{N_{1}+N_{2}} \rightarrow \mathscr{F}^{\prime}$ as,
\begin{align}
\varphi^{\prime}(\cdot)=\left[\begin{array}{c}{z(\cdot)} \\ {\varphi(\cdot)}\end{array}\right]
\label{eq:63}
\end{align}
and set $\omega^{\prime}=\left[\begin{array}{c}{\omega_{1}} \\ {\omega_{2}}\end{array}\right]$
\\Equation (64) can be described as,
\begin{align}
y(t)=\omega^{\prime T} \varphi^{\prime}(x(t))+b.
\label{eq:64}
\end{align}
In consideration of the fact that this concept is mathematically identical to the non-parametric form (53), all the numerical formulations are the same and thus the primary formulations for the semi-parametric approximation can easily be obtained with some minor adjustments.

First of all, $\Omega$ in (56) can be rewritten as,
\begin{align}
\begin{aligned} \Omega_{i j} &=<\varphi^{\prime}(x(i)), \varphi^{\prime}(x(j))>\\ &=\left[z^{T}(i), \varphi^{T}(x(i))\right]\left[\begin{array}{c}{z(j)} \\ {\varphi(x(j))}\end{array}\right] \\ &=z^{T}(i) z(j)+\varphi^{T}(x(i)) \varphi(x(j)) \\ &=z^{T}(i) z(j)+K(x(i), x(j)) .\end{aligned}
\label{eq:65}
\end{align}
Since it is identical to (65), the nonparametric form (66) can be transformed into,
\begin{align}
\begin{aligned} \hat{y}(t) &=\omega^{\prime T} \varphi^{\prime}(x(t))+b \\ &=\sum_{k=1}^{m} \lambda_{k} \varphi^{\prime T}(x(k)) \varphi^{\prime}(x(t)) \\ &=\sum_{k=1}^{m} \lambda_{k} z^{T}(k) z(t)+\sum_{k=1}^{m} \lambda_{k} K(x(k), x(t))+b, \end{aligned}
\label{eq:66}
\end{align}
and $\omega_{1}=\sum_{k=1}^{m} \lambda_{k} z(k)$. Therefore, (67) can also be written as,
\begin{align}
\hat{y}(t)=\omega_{1}^{T} z(t)+\sum_{k=1}^{m} \lambda_{k} K(x(k), x(t))+b.
\end{align}
From the description above, it can be shown that the nonlinear function $\phi(t)$ is mostly calculated as a kernel form,
\begin{align}
\phi(t)=\sum_{k=1}^{m} \lambda_{k} K(x(k), x(t)).
\end{align}

 It guarantees the computational sustainability of grey models with an undefined nonlinear function and it is quite simple to calculate the estimation of the general forms (49) and (50). For instance, the following basic notations for computational formulations can easily be obtained by using (50),
\begin{align}
\left\{\begin{aligned} y(k+1) &=X_{1}^{(1)}(k+1) \\ \omega_{1} &=\alpha \\ z(k) &=X_{1}^{(1)}(k) \\ \phi(k) &=\omega_{2}^{T} \varphi(\chi(k)) \\ \chi(k) &=\left[X_{2}^{(1)}(k), X_{3}^{(1)}(k), \ldots, X_{n}^{(1)}(k)\right]^{T}. \end{aligned}\right.
\end{align}
Comprehensive derivations can be contained in the earlier works of Xin Ma et al. \cite{ma2017novel, ma2018kernel, ma2016research, ma2016novel}.
\subsection{\textbf{Solutions of general formulations}}
Mathematically, the general formulation approaches use the same terminology as the linear formulations are as follows:
\subsubsection*{i. The continuous form of (49):}
\begin{align}
\hat{X}_{1}^{(1)}(t)=X_{1}^{(0)}(1) \cdot e^{-a(t-1)}+\int_{1}^{t} e^{-a(t-\tau)} \phi(\tau) d \tau
\end{align}
\subsubsection*{ii. Discrete form of (50):}
\begin{align}
\hat{X}_{1}^{(1)}(k+1)=X_{1}^{(0)}(1) \cdot \alpha^{k}+\sum_{\tau=2}^{k+1} \alpha^{(k+1-\tau)} \phi(\tau)
\end{align}
Remember that, only modified the linear function $f(\cdot)$ in (3) and (4) into $\phi(\cdot)$.

\section{Discussion}
\label{sec:dis}
Several promising Machine Learning models for long, medium, and short-term forecasting have been established using previous research findings data. It is noted that each of the mentioned strategies has a range of benefits and drawbacks. All have been carefully analysed and discussed in accordance to the study. Recently, the multivariate models gets particular consideration. The performance of the hybrid and ensemble models for time series forecasting was also proposed in recent review studies \cite{long2019novel,deb2017review,tealab2018time,hajirahimi2019hybrid,prilistya2020tourism,lara2021experimental,sanjay2017optimal}. Such forecasting models are essential for time series forecasting as well as effective optimization. To explain GML framework, the non-parametric approximation is the traditional Least Square Support Vector Machines (SL-LSSVM) suggested by Johan et al. \cite{suykens1999least}. This is one of the most common architectures for Machine Learning techniques. A semi-parametric approximation in the form of Slightly Linear Lowest Square Support Vector Machines (PL-LSSVM) proposed by Marcelo Espinoza et al. in 2004 \cite{espinoza2005kernel}, which has not been paid much more attention in recent years. In 2001, Bernhard et al., proved the Representer Theorem, which had very strong results and demonstrated the calculation still fits the form of (69) for any regularized formulation of traditional kernel approximation problems. These formulas are also valuable to design machines for the sake of applications \cite{scholkopf2001generalized}.

Furthermore, the linear differential in (49) and (50) is the partly understood dimension, similar to the general formulation of GML models poposed by \cite{Ma2018ABI} as described above. These differentials usually imply the state or performance sequence of models which are decreasing or increasing over time. Such mechanisms are also called dynamic systems which are widely used in real-world applications-including oil and gas prediction, weather forecasting, biomedical engineering, and so on. Whereas, the nonlinear function determined by the kernels represents a nonlinear aspect of the input sequence or the machine dependency series. These functions are clear and simply illustrate the nonlinear relationship between dynamic systems. As a result, GML models are nonlinear dynamic systems. Based on these attributes, the GML models were found to be much more efficient than the traditional linear grey models.

\subsection{\textbf{Future Perspectives}}
From the point of view of GML methodology, it can be demonstrated that the approaches of grey models and kernel-based methods can be successfully merged. The families of grey models and kernel-based models are very broad. However, we highlight some limitations of the GML framework addressed by the authors in their articles \cite{ma2018kernel,Ma2018ABI}. It may lead the researchers to enhance the GML concept efficiently, which are as under:

\subsubsection{\textbf{Based on Dataset}}
To discuss the modality of the dataset, GML framework specifically for the time series datasets. Therefore, it is not suitable for multimedia datasets or time-invariant situations. However, it is more efficient compared with other grey models for the big datasets, but still has some limitations of dataset size. Moreover, the GML framework is suitable for the single-time-line situation. Therefore, still there exist a need for improvement for multi-time-line datasets.

\subsubsection{\textbf{Based on Over-fitting}}
GML concept is a mixture of grey models and kernel-based models, ensuring that the shortcomings of such models can often be expressed in the applications. For example, LSSVM over-fitting happens in certain situations where the LSSVM has extremely high measuring performance but very low predicting accuracy. However, it does not happen with the GML, but the risk also exists. 

\subsubsection{\textbf{Based on Prediction}}
GML requires more data to yield more reliable results and ideal for short-term predictions, whereas LSSVM and linear grey models that in certain cases be useful for medium or long-term predictions. Research to solve these shortcomings should often be taken into consideration in prospective studies \cite{Ma2018ABI}.

Finally, it can be helpful to use these approaches to develop grey models or kernel-based models to enhance the GML framework in the future. Certain Machine Learning techniques can also be known to be useful for building GML models, including Multilayer Perceptrons (MP), Deep Learning Neural Networks, Gaussian Process Regression (GPR), etc. In addition to the range of other impressive Machine Learning, state-of-the-art approaches and the integration of such methods with grey models can also be useful in future research.

\section{\textbf{Conclusion}}
\label{sec:conclusion}
A comprehensive analysis of Machine Learning, Grey Forecasting Models, and GML was presented in this paper. Moreover, the increasing performance of the new combination models can be anticipated to be produced by the methods outlined in this article. Throughout our survey, a primer overview of Machine Learning was discussed based on different algorithms and applications by using big datasets. On the other hand, conventional grey models are specially developed for forecasting by using small time series datasets, which are divided into two types: Univariate and Multivariate models. From the traditional GM(1,1) model to the GML framework, each model is working for corresponding applications. Whereas, GML framework is derived from multivariate grey models. However, when there is some limited dataset of time series, GML has been demonstrated to outperform traditional LSSVM and grey models, which is the basic static framework, and showing the dynamic properties of GML models. The LSSVM analysis also indicated that the usage of established knowledge can significantly increase the efficiency of conventional Machine Learning models.

Developing new GML frameworks will increase more attention from different fields of scientists. More efficient GML models can be established in the future with the general formulation of GML. It is clear from this survey that GML not only develops many modern grey models but rather provides the possibilities of integrating the combination of traditional grey models as well as the dynamic nature of nonlinear Machine Learning models. Eventually, with the growing existence of a nonlinear dynamic system, more studies should also be carried out across a broader variety of real-world applications. Nonetheless, this survey illustrates the GML framework as well as the general perspective of Machine Learning and grey models, making it very useful for other applications as well as developing new GML models. This indicates that this framework allows researchers to use it for both theoretical and practical research.

\section*{Acknowledgment}
The authors would like to thank Dr. Xin Ma and Dr. Wenqing Wu at Southwest University of Science and Technology, China for their insightful suggestions and guidance of the theoretical and numerical analysis.

{
\scriptsize
\renewcommand\bibname{References}
\bibliographystyle{IEEEtran}
\bibliography{ref}
}

\EOD

\end{document}